\documentclass{ppig}
\usepackage{epsfig}
\usepackage{booktabs}
\usepackage{ucs}
\usepackage[utf8x]{inputenc}

\usepackage[]{blindtext}
\usepackage{dirtytalk}
\usepackage{url}
\usepackage{makecell}
\usepackage{tikzsymbols}

\usepackage[many]{tcolorbox}
\usepackage{xcolor}
\usepackage{varwidth}
\usepackage{environ}
\usepackage{xparse}
\usepackage{caption}
\usepackage{subcaption}
\usepackage{enumitem}

\title{Students' Feedback Requests and Interactions with the SCRIPT Chatbot:\\ Do They Get What They Ask For?}

\author{Andreas Scholl \\
  Computer Science \\
  Nuremberg Tech \\
  andreas.scholl@th-nuernberg.de \\
  \And
  Natalie Kiesler \\
  Computer Science \\
  Nuremberg Tech \\
  natalie.kiesler@th-nuernberg.de}

\date{}

\begin{document}
\maketitle
\thispagestyle{empty}

\begin{abstract}
Building on prior research on Generative AI (GenAI) and related tools for programming education, we developed SCRIPT, a chatbot based on ChatGPT-4o-mini, to support novice learners. SCRIPT allows for open-ended interactions and structured guidance through predefined prompts. We evaluated the tool via an experiment with 136 students from an introductory programming course at a large German university and analyzed how students interacted with SCRIPT while solving programming tasks with a focus on their feedback preferences. The results reveal that students' feedback requests seem to follow a specific sequence. Moreover, the chatbot responses aligned well with students' requested feedback types (in 75\%), and it adhered to the system prompt constraints. 
These insights inform the design of GenAI-based learning support systems and highlight challenges in balancing guidance and flexibility in AI-assisted tools. 
\end{abstract}

\section{Introduction}

Ever since the broad availability of GenAI, we have seen an increasing number of application contexts in computing education, and particularly introductory programming~\cite{prather2023wgfullreport,prather2025beyond}. 
This development is not surprising, as GenAI is capable of passing introductory programming exercises and courses~\cite{geng2023chatgpt,kiesler2023large,savelka2023large}. They can generate programming error messages~\cite{Leinonen2023,Sarsa2022,macneil2022experiences}, exercises~\cite{jacobs2025unlimited}, and formative feedback~\cite{Bengtsson_Kaliff_2023,kiesler2023exploring,jacobs2024evaluating,roest2023nextstep,lohr2025youre}. For example, it has been shown that it is possible to elicit certain types of feedback for programming novices by using specific prompts~\cite{lohr2025youre}.
Moreover, recent GenAI tools have advanced rapidly, and some of them can generate precise, structured, and personalized feedback for novice programmers~\cite{azaiz2024feedbackgeneration, phung2023generating, scholl2024hownovice}.

Due to GenAI's potential for learners, several prototypes and tools have been developed to integrate GenAI into programming education. 
Context-specific tools with feedback for novice learners of programming comprise, for example, CodeAid~\cite{kazemitabaar2024codeaid}, Codehelp ~\cite{liffiton2023codehelp}, LLM Hint Factory~\cite{xiao2024exploring}, and the StAP-tutor ~\cite{roest2023nextstep}. 
Other chatbots, such as the CS50 Duck~\cite{liu2024teaching,liu2025improving} and the CodeTutor~\cite{lyu2024evaluating} can provide context-aware feedback. 
Tutor Kai~\cite{jacobs2025unlimited} can generate comprehensive programming tasks, including problem descriptions, code skeletons, unit tests, and model solutions~\cite{jacobs2025unlimited} to help students practice and gain experience while also getting feedback~\cite{jacobs2025thatsnotthefeedback}. 
Generally speaking, all of these tools are designed to support novice programmers struggling with various challenges.
Among them are cognitively complex tasks in introductory programming (e.g., problem understanding, designing and writing algorithms, debugging, and understanding error messages~\cite{kiesler2024modeling,Luxton-Reilly2018,Ebert2016,spohrer1986novice,duboulay1986some}). 
Educators' high and unrealistic expectations~\cite{luxton-reilly2016,Luxton-Reilly2018,whalley2007many,kiesler_diss_2022}, a low student-educator ratio~\cite{petersen2016revisiting}, and an increasingly diverse student body in higher education (e.g., due to diverse learners and educational biographies) add to the list of novices' challenges. 

Considering the potential of GenAI and the challenging nature of introductory programming education, it is crucial to keep advancing tutorial systems capable of providing the feedback computing students need and want. Despite the availability of certain tools for novice programmers (e.g., CodeAid, etc.), there is no one-size-fits-all solution and no consensus on how to design a chatbot with pedagogical guardrails yet. 
Prior research analyzed students' interactions with GenAI tools, such as ChatGPT~\cite{scholl2024analyzing}. It revealed diverse usage patterns, e.g., superficial engagement, such as seeking quick solutions, but also extensive and iterative dialogues. The evaluation of the student perspective indicates their appreciation of GenAI's continuous availability and immediate responses~\cite{scholl2024hownovice}. At the same time, students expressed their concerns about over-reliance, inconsistent responses, and a lack of pedagogical guidance~\cite{scholl2024hownovice}. Motivated by these insights and in line with the lessons learned from the design of similar tools, we developed a \textit{Supportive Chatbot for Resolving Introductory Programming Tasks} (SCRIPT) for novice learners of programmers. It is based on ChatGPT-4o-mini, and integrates task description and context into the system prompt. Computing students are offered predefined prompts for specific feedback types~\cite{keuning2018,narciss2006} based on related work~\cite{lohr2025youre}. In addition, students can enter free-form input, and receive step-by-step responses~\cite{liffiton2023codehelp} without giving away (model) solutions.

In this research paper, the \textbf{goal} is to investigate students' interactions with SCRIPT in terms of requested feedback types. In addition, we evaluate to what extent the generated responses match students' requests and whether the output adheres to the system prompt constraints. The results \textbf{contribute} to increasing our understanding of GenAI-based chatbots for introductory programming education and how to design them to align with learners' needs. They particularly inform us of learners' needs in terms of AI-generated feedback for programming tasks. These findings thus have implications for tool developers and educators, but also computing students applying respective tools. 


\section{Related Work} 
\label{sec:relatedwork}

In the past few years, numerous application contexts were evaluated to show the potential of GenAI for computing education, and introductory programming in particular. For example, GenAI and related tools (e.g., ChatGPT, Codex, Copilot, or Llama) have been shown to successfully solve introductory programming tasks and pass respective exams~\cite{wermelinger2023using,geng2023chatgpt,kiesler2023large,savelka2023large}.
Other studies revealed they can enhance programming error messages~\cite{Leinonen2023}, and effectively generate code explanations~\cite{macneil2022experiences,Sarsa2022}. Explanations by GenAI were found to be capable of analyzing student code and providing instruction on how to fix errors~\cite{phung2023generating,zhang2022repairing}. Even elaborate feedback types were identified in GenAI output in response to student code~\cite{kiesler2023exploring,azaiz2024feedbackgeneration}.
In the following, we present relevant research on GenAI feedback, and customized GenAI tools that informed the development of our system (SCRIPT), and the research design. 

\subsection{Feedback Potential of GenAI}
The generation of feedback is considered a huge potential of GenAI, especially with the rapid advancements of the underlying large language models. A study by \citeauthor{phung2023generating} ~\cite{phung2023generating} investigated the use of LLMs to fix syntax errors in Python programs. They developed a technique to receive high-precision feedback from Codex. By using qualitative methods, several other studies explored the feedback characteristics of GenAI in response to student solutions containing errors~\cite{kiesler2023exploring,azaiz2024feedbackgeneration}. For example, \citeauthor{kiesler2023exploring} ~\cite{kiesler2023exploring} identified the following feedback elements: stylistic advice, textual explanations of the cause of errors and their fix, examples, meta-cognitive and motivational elements. However, they recognized misleading information and uncertainty in the model's output, depending on the task. ChatGPT-3.5 also requested more information in a few cases. 
A qualitative evaluation of GPT4 Turbo's feedback showed notable improvements~\cite{azaiz2024feedbackgeneration}. The outputs were more structured, consistent, and always personalized~\cite{azaiz2024feedbackgeneration}.


\citeauthor{lohr2025youre}~\cite{lohr2025youre} investigated whether they can generate specific feedback for introductory programming tasks using GenAI (i.e., ChatGPT). Following an iterative approach, they designed prompts to elicit specific feedback types, such as knowledge about mistakes, or knowledge on how to proceed (cf. ~\cite{narciss2008feedback,keuning2018}). They qualitatively evaluated the generated output with human intelligence and determined its feedback type to check if the prompts had elicited the expected feedback. As a result, they present prompts for the generation of different types of feedback suitable for introductory programming tasks and student submissions~\cite{lohr2025youre}. They also noted that misleading information occurred less frequently compared to related work~\cite{kiesler2023exploring} \cite{lohr2025youre}.

\subsection{Customized GenAI Tools for Programming Education}
Due to their feedback potential, GenAI models are being integrated into educational tools and systems to offer novice-friendly explanations tailored to individual student errors. The dcc -\,-help tool, for example, integrates ChatGPT 3.5 into the Debugging C Compiler (DCC). It produces context-aware explanations in response to compiler errors in DCC~\cite{taylor2024dcchelperrorexplanations}. A more recent conversational AI extension to the GenAI-enhanced C/C++ compiler DCC is DCC Sidekick~\cite{renzella2025compiler}. It generates pedagogical programming error explanations without integrating student input, and by applying the Socratic method.

Another example of a programming assistant is CodeAid~\cite{kazemitabaar2024codeaid}. Its implementation avoids the generation of code solutions to support students' thinking and learning. CodeAid offers help with general questions, inline code explorations, questions from code, help to fix code, code explanations, and help in writing code. Yet, the developers did not distinguish or implement any well-known and context-specific feedback typology (cf.~\cite{keuning2018}). \citeauthor{kazemitabaar2024codeaid} (2024) summarize four design considerations for AI coding assistants: (1) exploiting unique advantages of AI, (2) designing the AI querying interface, (3) balancing the directness of AI responses, and (4) supporting trust, transparency and control.

CodeHelp is another GenAI-powered tool designed to provide guardrails and on-demand programming assistance~\cite{liffiton2023codehelp}. Its features comprise a simple interface for students' help requests, a system prompt avoiding the generation of complete code solutions, and a sufficiency check to catch ambiguous or incomplete student queries. CodeHelp was evaluated by 52 students over a 12-week period. The results show students value the availability and help of the tool, and that it complements the efforts of the instructor~\cite{liffiton2023codehelp}.

Xiao et al. (2024) followed a somewhat different approach by investigating whether and how different hints support students' problem-solving and learning processes in the LLM Hint Factory~\cite{xiao2024exploring}. The latter is a system providing four levels of hints, from natural language to concrete code assistance. Via a think-aloud study with 12 programming novices, they identified recommendations for feedback design. For example, hints about the next step or how to solve syntax issues should be concise and personalized to students' requests to meet their needs. Otherwise, students may switch to ChatGPT to receive full solutions according to the authors of the study~\cite{xiao2024exploring}.
Similarly, Roest et al. 2023 developed the StAP-tutor; a GenAI-based tutoring system for next-step hints~\cite{roest2023nextstep}. The system was designed by exploring various prompts and evaluating the feedback. The best output was generated for inputs with a problem description and the words \say{student} and \say{hint}. Another recommendation is to increase the temperature parameters of the used model. 
Through a student and expert evaluation, the feedback was evaluated as concise and personalized, but it still contained misleading information for OpenAI's GPT-3.5-turbo model. Thus, future work is required to evaluate more recent models~\cite{roest2023nextstep}.

There are some other recent applications and studies on the use of GenAI in computing. For example, Yeh et al. (2025) investigated an interactive approach to teach students how to write better prompts so they can eventually generate working code~\cite{yeh2025bridging}. Bhowmick and Li (2025) experimented with LANTERN to teach students relational query processing as part of a database course ~\cite{bhowmick2025experience}. 
Similarly, Riazi and Rooshenas (2025) explored GenAI feedback to support students' conceptual design competencies (e.g., via entity-relationship diagrams) in a database systems course~\cite{riazi2025llm}.
Forden et al. (2025) developed an automated assessment tool trying to cultivate students' coding style and foster timely submissions~\cite{forden2025unlocking}. A last example is SENSAI, an AI-powered tutoring system for teaching cybersecurity~\cite{nelson2025sensai}. It utilizes the learner's workspace, i.e., active terminals and edited files as input, highlighting the importance of context in the system prompt.

\subsection{Research Gap}
It is crucial to keep investigating newly emerging large language models (e.g., ChatGPT 4o, 4.5, and OpenAI o1, o3), which is due to the fast pace of this technology~\cite{prather2023wgfullreport}. 
This is particularly relevant as research data (e.g., for benchmarking) are usually not available~\cite{kiesler2023why,prather2023wgfullreport}. Few studies utilize the same dataset for evaluating or benchmarking recent models. 
Moreover, the general limitations of GenAI models are well-known. Among them are inaccuracies, hallucinations, misleading information (especially for novice learners), data and privacy concerns, reproduction of bias and stereotypes, lack of accessibility, and their in-transparency by design ~\cite{gill2024transformative,prather2023wgfullreport,zhai2022chatgpt,kiesler2023exploring,kiesler2025role,alshaigy2024forgotten}. Hence, we need to critically reflect upon the use of GenAI, and how to provide pedagogical instruction and guardrails for students. The same applies to the development and design of tools based on GenAI models. 
A major challenge in this design is finding a balance between individual student support and avoiding the extensive generation of model solutions. This can be addressed by restricting the chatbot from generating any code at all (cf. ~\cite{liffiton2023codehelp}). 

Our goal, however, is to enable more flexible interactions, where the chatbot can produce helpful code snippets without giving away complete solutions. We also postulate that students can benefit from individual feedback and help, and code snippets may be helpful for some of them. In addition, we want to provide exemplary prompts leading to specific types of feedback, e.g., indicating and explaining error(s), or the next steps to solve them, as successful prompting can be challenging for novices as well. 
Moreover, we need to integrate students into the discussion to qualitatively evaluate how students apply GenAI tools (customized or not). 
Therefore, a critical aspect of this work is whether AI-generated feedback addresses students' informational needs. 


\section{Methodology} 
\label{sec:meth}
In this study, we present SCRIPT, a chatbot based on ChatGPT-4o-mini. It is designed to support novice learners of programming seeking feedback. The evaluation of SCRIPT is guided by the following research questions:
\begin{enumerate}[leftmargin=13mm] 
\itemsep0.5pt
    \item[RQ1]\textit{How do students interact with SCRIPT in the context of introductory programming tasks?}
    \item[RQ2]\textit{To what extent does the generated output match students' requests?}

    \item[RQ3]\textit{To what extent do the generated outputs adhere to the system prompt constraints?}
%


\end{enumerate}

To address these RQs, we utilize empirical data from students (n=136) enrolled in an introductory programming class at a large German university. Students were asked to solve programming exercises at home while having access to the SCRIPT chatbot. The chat protocols were automatically recorded, and students were asked to rate the GenAI-powered responses and leave comments (all voluntarily). 
In the following subsections, we introduce SCRIPT, the course context, selected tasks, and data analysis methods. Our research data (e.g., tasks, system prompt, interactions, etc.) are available in an online repository~\cite{Scholl_Kiesler_2025_datastudents}.

\subsection{Introducing SCRIPT}
\begin{figure}[htb]
\centering
\includegraphics[width=\linewidth]{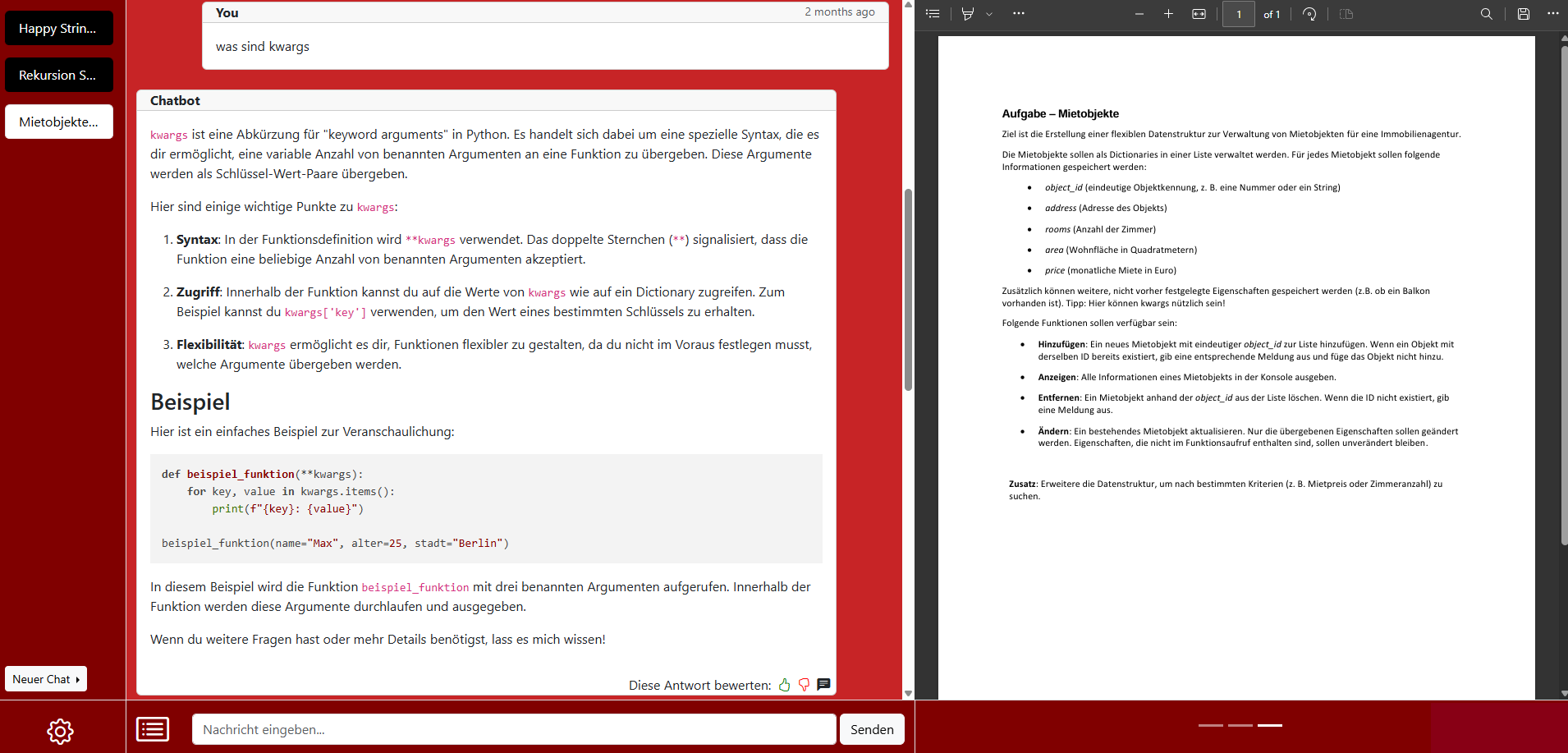}
    \caption{Overview of the SCRIPT User Interface}
    \label{fig:script}
\end{figure}

SCRIPT~\cite{scholl2025script} is implemented as a standalone web application, which is accessible via a link in the respective university's Moodle course (no additional sign-in needed). 
The setup enables anonymous data gathering (via anonymous IDs). The user interface (UI) is depicted in Figure \ref{fig:script}. It is similar to the structure of ChatGPT's interface. Each chat session is displayed at the very left of the UI. It is task-specific, meaning a conversation is dedicated to solving one programming problem. Students can initiate multiple conversations, including revisiting the same task. The chatbot receives both the task description (displayed on the right) and a reference solution as context (i.e., as part of the system prompt~\cite{Scholl_Kiesler_2025_datastudents}), eliminating the need for students to manually copy and paste these elements into the chat, which is supposed to ease its use~\cite{kazemitabaar2024codeaid}. Students can rate each GenAI response with a thumbs-up or down feedback, and via textual comments (all optional).
The backend architecture relies on a Node.js environment with an Express/Socket.io server. The data management is handled via an SQL database (MariaDB). Student interactions and their feedback (thumbs-up/down ratings, and open input) are logged.

SCRIPT provides two distinct prompting options, which are designed to exploit the unique advantages of GenAI while balancing guidance and flexibility~\cite{kazemitabaar2024codeaid,yeh2025bridging}. \textit{Closed prompts} are predefined based on recent research on AI-generated feedback ~\cite{lohr2025youre} and an established feedback typology for programming contexts~\cite{keuning2018,narciss2006}. These closed prompts guide students through problem-solving steps and issues, including understanding task constraints (KTC), identifying required programming concepts (KC), recognizing mistakes (KM), determining how to proceed (KH), assessing performance levels (KP), and evaluating code correctness relative to a reference solution (KR). These closed prompts are available for students next to the input field. They are supposed to encourage students' structured engagement with the chatbot, and reduce the challenges of starting with the problem-solving process, or formulating a prompt (cf. ~\cite{scholl2024hownovice,scholl2024analyzing}). 

\textit{Open prompts} are common when using a chatbot, and are comparable to those expected by CodeHelp~\cite{liffiton2023codehelp}. They denote students' free-form textual input. When students enter text, SCRIPT prevents the generation of complete code as output. Instead, it fosters step-by-step problem-solving by identifying errors in student code without directly correcting them. Thus, the tool provides stepwise hints and no complete solutions in the form of code. It also offers examples (unrelated to the specific task) and templates (incomplete code structures with only comments in key sections). SCRIPT is available at any time, so students can work at their own pace, with all tasks being accessible. This is supposed to support mastery learning where learners can repeatedly engage with a task until achieving proficiency~\cite{szabo2025modelsofmastery,keller1968good}. The dual-mode interaction structure with closed and open prompts allows students to choose between guided problem-solving and a more exploratory learning approach.
The system prompts for both modes are available in the online repository~\cite{Scholl_Kiesler_2025_datastudents}.

\subsection{Course Context and Task Selection}
SCRIPT was designed to support students in an introductory programming course for first-year computing students ($N=666$) at Goethe University Frankfurt (Germany). The present study was conducted in the winter term 2024/25. The majority of students were enrolled in the bachelor's degree program in Computer Science (CS). Only some were enrolled in other disciplines or chose CS as a minor. Prior programming experience was not required to participate. A Moodle course provided access to learning materials, including SCRIPT. The course structure included a weekly two-hour lecture for all students. Tutorial sessions (20 students each) accompanied the course. Tutorial sessions typically last two hours every week. A key component of the tutorials is the weekly or biweekly exercises, so students can gain hands-on programming experience. Students work individually on tasks and earn exercise points for their submissions. These contribute to the final exam score.

For this study, we designed three programming tasks for students to be completed using SCRIPT. The tasks were supposed to be solved within one week, starting on January 13, 2025. Students had the opportunity to engage with one or more tasks. Students were not instructed on how to use SCRIPT. Participation was voluntary, but students could earn four exercise points as an incentive. Furthermore, the purpose of the study and its procedures were introduced to students during the lecture preceding their tutorial sessions.
Table \ref{tab:task-selection} summarizes the tasks used in this study and the concepts they address. The given tasks were aligned with the course curriculum and the student's assumed/expected level of expertise. The course's facilitator also ensured that the tasks were relevant and adequate. The full task descriptions are available along with the other research data~\cite{Scholl_Kiesler_2025_datastudents}.  

\begin{table}[htb]
\scriptsize
\centering
\begin{tabular}{l|l|c}
 \textbf{Task} & \textbf{Description} & \textbf{Concepts}\\
\hline
 \textit{Happy Strings} & \makecell[l]{
 Compute the number of \say{happy} strings within all 
 sub-strings of a given string\\of digits. A string is considered \say{happy} if the digits can be rearranged to\\repeat twice. The following steps are required: (a) Test for \say{happy} property;\\(b) iterate and test all substrings.} & \makecell[t]{recursion, functions, \\lists, conditionals, \\string manipulation} \\
 \cline{1-2}
 \makecell[l]{\textit{Recursion} \\\textit{Snippets}} & \makecell[l]{Determine recursion type, return value and number of function calls of 4\\functions: (a) sum of digits; (b) list reversal; (c) multiplication\\(d) Ackermann function.} &  \\
 \hline
 \makecell[l]{\textit{Rental} \\\textit{Properties}}
 & \makecell[l]{
 Implement a data structure to manage rental properties  
 using dictionaries in\\a list. Provide functions to add, display, remove, and update properties based\\on unique 
 IDs. Extend functionality to allow searching by criteria.} & \makecell{functions, lists, \\dictionaries, \\keyword parameters, \\data structures}
\end{tabular}
\caption{Selected tasks with short description and required programming concepts.}
\label{tab:task-selection}
\end{table}

\subsection{Data Analysis}
To examine students' interactions with SCRIPT, we collected and analyzed chat sessions from 136 students with their input and SCRIPT's output. The chat logs were stored as individual sessions (without any personal data or identifier). They also include students' ratings of SCRIPTs' responses, and textual feedback (if any). The chat logs constitute the data basis for RQ1, RQ2, and RQ3. 

For RQ1 and RQ2, we particularly focused on the student prompts and SCRIPTs responses to them. All of these questions and answers were analyzed and categorized regarding the requested and generated feedback type(s):  knowledge of result (KR), knowledge of correct result (KCR), knowledge of performance (KP), knowledge about task constraints (KTC), knowledge about concepts (KC), knowledge about mistakes (KM), knowledge on how to proceed (KH), and knowledge about meta-cognition (KMC)~\cite{keuning2018}.
The analysis followed a structured coding approach. We started with approximately 15\% of the chat sessions to confirm the adequacy of the deductive coding scheme based on the literature~\cite{keuning2018}. It applied to almost all feedback requests and responses. However, some additional requests and responses were noticed and coded inductively based on the material.  
For example, students' prompts were categorized as: technical (TEC), social interaction (SoI), answer to GenAI question (ANS), clarification question (WHAT), off-topic (OFT), new task request (TR), incomprehensive (IN), and prompt injection (HACK).
Similarly, SCRIPTs' responses were coded with these additional categories: marked as offensive (OFF), denied request (DENY), social (SoI), technical (TEC), off-topic (OFT), technical error (TE), and new task (TR).
(Definitions and anchor examples of the additional categories are  available in the data publication~\cite{Scholl_Kiesler_2025_datastudents}.)
As part of this rule-guided coding process, multiple codes (a maximum of 3) were applied to a coding unit (i.e., a student input and a generated response each was considered a single coding unit). A student's entire chat session was considered a context unit, in case of uncertainties (e.g., regarding the intentions of a student trying to elicit feedback or have a nice chat). 
After the initial round, deductive and inductive categories were applied to the remaining material. All of the collected chat data were coded twice by the same coder (first author of this work) to ensure internal consistency and reliability. The coder had prior experience in qualitative coding.

To answer RQ1, the student requests and chatbot responses were aggregated and processed into a flowchart to represent the interactions and their frequencies. 
Regarding RQ2, we compared the feedback type request of every student question with the feedback category/categories evident in SCRIPTs' responses (both in an aggregated and pair-wise form). We present the number of matches, over-matches (i.e., a match of requested and generated feedback types, plus additional feedback categories in the response), partial matches, or mismatches.  


The last research question (RQ3) had the goal of evaluating the quality of SCRIPTs' outputs in terms of correctness, and step-wise hints. We addressed those aspects through the following indicators (cf.~\cite{azaiz2024feedbackgeneration,azaiz2025opensmallrigmarole,roest2023nextstep}): (1) \textit{Number of problem-solving steps} SCRIPT provides in a response. Per system prompt, a single step per response should be provided. (2) \textit{Solution} provided to the task (partial or complete), whereas no full solution should be generated. (3) \textit{Code examples}; SCRIPT is instructed to provide only simple examples. (4) \textit{Code templates}; SCRIPT may provide code templates to students, consisting of structural frameworks for a function, loop, or conditional headers. (5) \textit{Code Corrections}; SCRIPT should correct students' code by pointing out the mistake and providing hints. Hence, we evaluate whether SCRIPT provides corrected code. (6) \textit{Correctness of Response}; we determine if a response from SCRIPT is correct, partially correct, or incorrect.

\section{Results} 

In total, 136 students engaged with SCRIPT, generating 241 chat sessions. Across these interactions, students submitted 1,409 prompts, and SCRIPT generated 1,409 responses. The distribution of prompts per student varied, though, with a mean of 10.36 (SD = 10.86) and a median of 7. 
Among the student prompts were 207 predefined \say{closed} prompts, requesting the following feedback types: KC (54), KTC (48), KR (49), KM (26), KP (17), and KH (13).

\subsection{Students' Interactions with SCRIPT (RQ1)} 
\begin{figure}[htb]
\includegraphics[width=\linewidth]{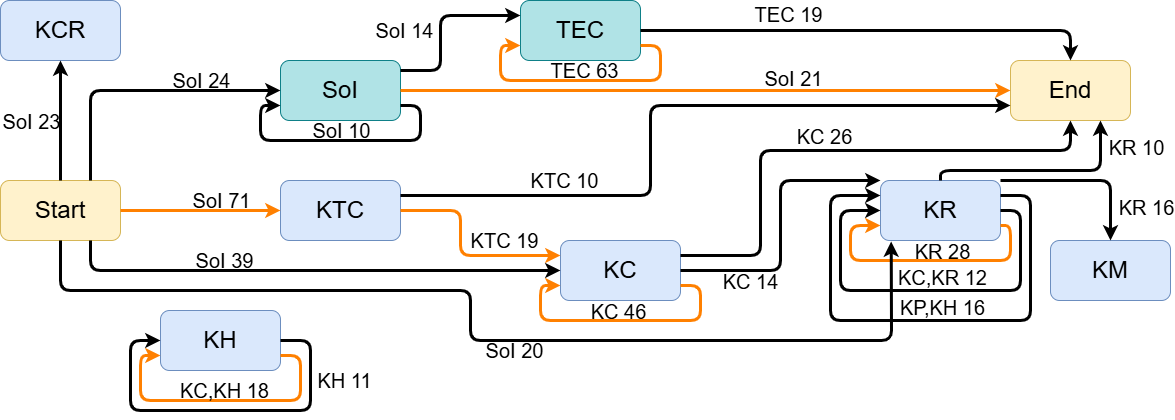}
\caption{Flowchart of students' interactions. Nodes mark students' feedback requests, edges represent SCRIPTs' feedback categories. Only interactions that occurred at least 10 times are represented.}
\label{fig:flowchat}
\end{figure}

To answer RQ1, we analyzed the feedback requests and generated responses w.r.t. their feedback types. The resulting feedback (and other) categories of inputs and outputs were aggregated in a flowchart, as displayed in Figure \ref{fig:flowchat}. It illustrates common interaction patterns between students and SCRIPT. Nodes indicate the feedback types of students' prompts, and edges represent the feedback generated by SCRIPT. The flowchart includes loops, such as for KH, which illustrate repeated requests for the same feedback type. Next to the feedback type, we always provide the number of occurrences. Figure \ref{fig:flowchat} only represents those interaction patterns observed at least ten times, and thus were more common. It is important to note that the figure represents the results in an aggregated form and does not display individual student paths. A more detailed flowchart and the raw data are available online~\cite{Scholl_Kiesler_2025_datastudents}.

In the following, we describe the most frequent requests and responses as depicted in Figure \ref{fig:flowchat}. 
After a default greeting by SCRIPT, students began their session by asking questions about the task constraints (KTC) in 71 cases. This was often followed by questions (19) regarding the necessary programming concepts (KC). Next, students (46) engaged in multiple follow-up inquiries, requesting further explanations or clarifications of concepts. Then they submitted their (partial) solution and asked SCRIPT to evaluate its correctness (KR, 14). This process often occurred iteratively, with students refining their code based on SCRIPT's feedback (28). Instead of simply providing a binary correct/incorrect response, SCRIPT typically offered reasoning about the student's performance (KP, 16), clarifying key concepts (KC, 12), or suggesting how to approach corrections (KH, 16). However, when students used predefined \textit{closed prompts}, SCRIPT adhered to a stricter response format, delivering only KR feedback (i.e., correct/incorrect). This often led students to seek additional clarification about their mistakes through \textit{open prompts}.
Another interaction pattern was students requesting the correct solution (KCR) at the beginning of the session. In 18 of the 23 cases, SCRIPT explicitly denied these solution requests. 

In 24 chat sessions, students initiated conversations with social interactions, sometimes (10) followed by additional social exchanges. In 14 sessions, this led students to ask technical questions about SCRIPT before engaging with the task. Some sessions (19) ended at this step without students attempting to solve the problem.
Regardless of how a student started or concluded their chat, students asked how to proceed (KH) at different stages of the problem-solving process (see Figure \ref{fig:flowchat}). KH loops have occurred at least 10 times (KH, 11; KC, KH, 18). Thus, the number of outgoing requests from KH to other feedback types was less than 10, which is why the KH node is isolated in the aggregated flowchart.

SCRIPT typically responded with step-by-step explanations, often adding conceptual knowledge (e.g., KC, KH 18) to guide students through the task and next steps. This iterative exchange generally continued until the student reached a solution. However, a few students relied solely on requesting the next steps without adding their code as input. In such cases, SCRIPT constructed the solution incrementally, guiding the student through each step of the process.

\subsection{Matches of Student Requests and SCRIPT's Responses (RQ2)} 
To understand the extent to which the generated output matched students' requests, we first of all aggregated all requested and provided feedback types in Table \ref{tab:rq2-feedback-types}. It shows the total number of occurrences of requested and generated feedback types across all student prompts and responses. For most feedback categories, the numbers align closely, indicating that SCRIPT generally generated what students requested. 
For some requests, SCRIPT supplemented its responses with additional conceptual explanations (KC, 111) and guidance on how to proceed (KH, 106), particularly when students asked for feedback for their code (i.e., had requested KM (27), or KR (37)). For this reason, the number of feedback types in the responses exceeds the number of requests for KC, and KH (see Table \ref{tab:rq2-feedback-types}). 






\begin{table}[h]
\begin{subtable}[h]{0.45\textwidth}
\scriptsize
\centering
\begin{tabular}{c|c|c}
 Feedback Type & No. Requests & No. Responses\\
 \hline
 KTC & 158 & 147\\
 KC & 295 & 424\\
 KH & 159 & 352\\
 KM & 84 & 81\\
 KP & 36 & 71\\
 KR & 209 & 149\\
 KCR & 100 & 120\\
 KMC & 6 & 6\\
\end{tabular}
\caption{Feedback Types}
\label{tab:rq2-feedback-types}
\end{subtable}
\hfill
\begin{subtable}[h]{0.45\textwidth}
\scriptsize
\centering
\begin{tabular}{c|c|c}
 Additional Cat. & No. in Prompts & No. in Responses\\
 \hline
 TEC & 165 & 152\\
 SoI & 101 & 106\\
 ANS & 141 & -\\
 WHAT & 71 & -\\
 OFT & 48 & 45\\
 TR & 15 & 15\\
 IN & 14 & -\\
 HACK & 26 & -\\
 DENY & - & 42\\
 OFF & - & 9\\
 TE & - & 58\\
\end{tabular}
\caption{Additional Categories}
\label{tab:rq2-other-types}
\end{subtable}
\caption{Total number of requested and generated (a) feedback types and (b) additional categories (both not pair-wise); 
prompts and responses may contain multiple types each.}
\end{table}

We also evaluated to what extent the student feedback requests and the immediate responses from SCRIPT were aligned (i.e., in the question-response pairs). It should be noted though that not all student inputs requested feedback, which is why we only evaluated 891 question-response-pairs. In 47\% of cases (417 out of 891), the chatbot's feedback directly matched what students requested. 
We also observed so-called ``over-matching''. That is, SCRIPT provided additional feedback types, and not just those that were explicitly requested.
Considering these, the total alignment increased to 75\%, i.e., an additional 255 out of the 891 pairs were over-matching. Overmatching primarily occurred for KC and KH categories, where SCRIPT offered extra explanations or next-step guidance beyond what was explicitly requested.
In 22\% of question-response pairs (195 out of 891), the response did not match the intended feedback type (mismatch). Notably, 25\% of these mismatches (49 out of 195) involved students requesting direct solutions (KCR), which SCRIPT was designed to reject. This indicates that while mismatches occurred, a quarter resulted from SCRIPT correctly adhering to its constraints -- rather than failing to provide the desired feedback. 58 technical errors (TE) occurred when SCRIPT refused to answer requests for KTC and KC feedback, due to missing student code (KTC and KC did not require student code to be generated, though). 

Students themselves provided 151 thumbs-ratings (130 up, 21 down) and 29 short written comments. Positive remarks highlighted SCRIPT's clear explanations and helpful guidance (\say{Providing a template without giving everything away is helpful}). Critical comments mostly addressed incorrect evaluations, overly confident, or brief responses (\say{Here, the AI was confidently wrong.}).


\subsection{SCRIPT's Adherence to System Prompt Constraints (RQ3)} 
To evaluate SCRIPT's adherence to the system prompt constraints, we evaluated (1) the number of problem-solving steps, the generation of (2) solutions, (3) code examples, (4) templates, (5) code corrections, and (6) the correctness of responses of all 1409 generated responses (see Table \ref{tab:rq3-prompt-constraints}). After restricting the responses from SCRIPT to problem-related answers (removing the categories TEC, SoI, DENY, OFF, OFT, TE, and TR), 1016 responses remained for analysis.

(1) In 51\%, SCRIPT adhered to the constraint of providing only one step at a time in the problem-solving process. 21\% of the responses comprised an overview containing multiple steps but included a clear starting point or next step (see Table \ref{tab:rq3-prompt-constraints}).
(2) Across all tasks, students made 100 KCR requests. SCRIPT provided solutions in 42 cases. These were usually granted after a stepwise generation of problem-solving steps. The success rate of solution requests varied across tasks, with \say{Recursion Snippets} having the highest proportion of fulfilled requests (71\%), followed by \say{Happy Strings} (38\%) and \say{Rental Properties} (25\%).
(3) Simple code examples were generated 104 times, and complex examples were observed in 3 responses. 
(4) Regarding template generation, SCRIPT correctly provided 167 templates in the intended format, offering general code structures without implementations. However, in 95 responses, it eventually provided completed templates, after progressively building up the solution and providing several partial templates. 
(5) SCRIPT corrected the code of students 34 times and displayed it, despite the constraint not to do so.
(6) Overall, the correctness of SCRIPT's responses was high, with 81\% of answers being fully correct. 14\% of responses were at least partially correct, leaving only 5\% of answers classified as completely incorrect.




\begin{table}[htb]
\scriptsize
\centering
\begin{tabular}{c|c||c|c||c|c||c|c}
 \multicolumn{4}{c||}{(1) Number of solving steps in response} & \multicolumn{4}{c}{(6) Correctness of response}\\
 \hline
 \multicolumn{2}{c|}{Single Step} & \multicolumn{1}{c|}{514} & \multicolumn{1}{c||}{51\%} & \multicolumn{2}{c|}{Correct} & \multicolumn{1}{c|}{822} & 81\% \\
 \hline
 \multicolumn{2}{c|}{Multiple Steps} & \multicolumn{1}{c|}{289} & \multicolumn{1}{c||}{28\%} & \multicolumn{2}{c|}{Partially Correct} & \multicolumn{1}{c|}{141} & 14\% \\
 \hline
 \multicolumn{2}{c|}{Multiple Steps, Explicit Next-Step} & \multicolumn{1}{c|}{213} & \multicolumn{1}{c||}{21\%} & \multicolumn{2}{c|}{Incorrect} & \multicolumn{1}{c|}{53} & 5\%\\
 \hline
 \multicolumn{8}{c}{}\\
 \multicolumn{2}{c||}{(2) Solution given} & \multicolumn{2}{c||}{(3) Given Examples} & \multicolumn{2}{c||}{(4) Code templates} & \multicolumn{2}{c}{(5) Code corrections}\\
 \hline
 Partial & 102 & Simple & 104 & Provided & 167 & Corrected & 34 \\
 \hline
 Complete & 129 & Complex & 3 & Completed & 95 & \multicolumn{2}{c}{} \\
 \cline{1-6}
\end{tabular}
\caption{Indicators for SCRIPT's adherence to the system prompt constraints (for 1016 responses).}
\label{tab:rq3-prompt-constraints}
\end{table}

\section{Discussion} 
Our findings revealed interaction patterns of introductory programming students using SCRIPT. When using the chatbot, students seemed to follow a certain (problem-solving) sequence, that has not been revealed in previous research: understanding the task constraints (KTC), identifying relevant programming concepts (KC), formulating a solution approach (KH), debugging errors (KM), verifying results (KP, KR), and ultimately comparing with the correct solution (KCR). While most interactions focused on problem-solving, some students also engaged in technical inquiries (TEC) and social exchanges, highlighting a broader spectrum of engagement beyond the mere task. Notably, KH inquiries and follow-up questions from the chatbot led to more student interactions (i.e., follow-up questions), with SCRIPT guiding students step by step. 

Related to RQ2, the chatbot's responses aligned well with students’ needs in most cases, particularly in clarifying task constraints (KTC), which were addressed with good accuracy. Concept explanations (KC) were generally good but occasionally too complex or lacking specific details. This suggests future improvements of the chatbot in tailoring explanations more precisely to the task. Performance feedback (KP) showed a high degree of variation and seemed random. KR feedback in \textit{closed prompts} was mostly binary, so students usually continued with additional \textit{open prompts} to elicit KM feedback. This may indicate that students perceived it as insufficient or had different expectations towards the generated response. Solution requests (KCR) resulted in the highest mismatch rate. This was expected as the chatbot was designed to reject such queries. Respective students tried to circumvent this via step-wise questions and gradually reconstructing the solution.

While SCRIPT generally adhered to system prompt constraints, some inconsistencies were observed. Especially for the \textit{Recursion Snippets} task, we found that solutions were often provided too early.
Code examples were consistently brief and simple. Code templates initially adhered to the guidelines but tended to be complete after several student inquiries. Direct solution requests were blocked as expected. Yet, in step-wise interactions, partial solutions emerged sooner or later, depending on the task. From a pedagogical perspective, this may not necessarily be a problem -- as long as students actively process the feedback. In general, it seems recommendable to balance direct AI responses~\cite{kazemitabaar2024codeaid}, and prevent the generation of complete code solutions~\cite{liffiton2023codehelp}. Importantly, SCRIPT demonstrated robustness against prompt injection attempts, successfully resisting nearly all manipulation efforts, except for one.

An additional observation concerns the chatbot's responses to \textit{closed prompts} (based on~\cite{lohr2025youre}), which were more formal, often writing in the third-person. In contrast, students' \textit{open} inputs exhibited a more conversational tone. We perceived \textit{closed} and \textit{open prompts} as two distinct conversational modes, with some abrupt transitions between the two. This observation may be attributed to the varying perspectives SCRIPT is required to adopt: \textit{closed prompts} request structured, restricted guidance, while \textit{open prompts} may shift the chatbot's role to a personal assistant -- depending on the student input. 

In general, the \textit{closed prompts} from prior work seemed useful~\cite{lohr2025youre}, which is reflected in students' use of them (207 times) and the general match between student requests and generated responses. It may also be helpful for students to get started or formulate a specific prompt, e.g., due to language barriers, anxiety, or insecurity about how to use technical terms. 
At the same time, it seems advisable to offer users multiple options, i.e., both \textit{closed} and \textit{open prompts}, thereby bridging guidance and flexibility, as suggested in related work~\cite{yeh2025bridging}.

Finally, we noted that the feedback typology from prior work~\cite{keuning2018} was sufficient to describe the requested and generated feedback items. 
This is worth discussing, as it was constructed based on the existing learning environment at the time, not chatbots and/or GenAI.
However, due to the new presentation (i.e., modality, adaptation~\cite{narciss2006}) of the feedback via the chatbot, we identified additional categories, such as social interactions (SoI), and many others. As GenAI and related tools advance rapidly, we expect to see new feedback categories, or changed presentation modes, etc., in the near future.


\subsection{Limitations}
Some of the study's limitations should be noted. For example, students knew their interactions were being analyzed, potentially influencing their behavior~\cite{roethlisberger1939management}. SCRIPT was also 
used in an unsupervised setting, allowing students to engage freely. Few students experimented with the tool, using social or technical prompts without attempting to solve the task, which may have affected interaction patterns. 
The study also relies on a researcher-driven analysis. Thus, students were not explicitly asked to categorize their feedback types. They might have had slightly other intentions or informational needs than those resembled in their prompts. Yet, we tried to mitigate this limitation by allowing student feedback (via thumbs, and open input field) to each generated response.
Finally, this study was conducted in a single university course in one country, limiting the generalizability of the findings. While the number of participants and responses strengthens its representativeness~\cite{boddy2016sample}, results may not be transferable to different curricula or institutions.

\section{Conclusions and Future Work} 

In this study, we investigated students’ interactions with SCRIPT, a GenAI-based chatbot developed to support problem-solving in introductory programming education. The evaluation involved 136 students, who solved dedicated programming tasks using SCRIPT in an unsupervised, self-paced setting. Chat sessions were analyzed regarding students' interactions and feedback requests, prompt-response alignment, and the system's adherence to constraints. Our analysis showed that students seemed to follow a certain sequence of feedback requests: KTC, KC, KH, KM, KP and KR, and, finally, KCR. Overall, SCRIPT provided correct responses in alignment with students' requests in 75\% of its responses.
Moreover, the system prompt was suitable for guiding students through step-by-step problem-solving without revealing full solutions. 
The chatbot remained robust and followed the instructional design.

The study's findings can inform the design of better prompts and scaffolded feedback sequences to help students work more independently with such tools. Future work should explore the use of GenAI tools with a broader range of task types and learning contexts, and continuously improve these early tools and prompts. In particular, closing the gap between \textit{open} and \textit{closed prompts} remains a key challenge, as it requires balancing structure and flexibility while managing the shift between the educator's role of restricting responses and supporting student exploration. This will be one of the next steps in advancing SCRIPT, to make the conversation flow more naturally. Moreover, we will use the students' feedback to improve the technical implementation, User Interface, and interaction. Continuing this work will help support novice programmers seeking feedback by SCRIPT. We also encourage other CS researchers, educators, and tool developers to keep exploring educational tools so we can leverage the potential of GenAI for good.  

\newpage 

\bibliography{ppig-sample-bibliography}

\begin{thebibliography}{}

\bibitem [\protect \citeauthoryear {%
Alshaigy%
\ \BBA {} Grande%
}{%
Alshaigy%
\ \BBA {} Grande%
}{%
{\protect \APACyear {2024}}%
}]{%
alshaigy2024forgotten}
\APACinsertmetastar {%
alshaigy2024forgotten}%
\begin{APACrefauthors}%
Alshaigy, B.%
\BCBT {}\ \BBA {} Grande, V.%
\end{APACrefauthors}%
\unskip\
\newblock
\APACrefYearMonthDay{2024}{}{}.
\newblock
{\BBOQ}\APACrefatitle {Forgotten Again: Addressing Accessibility Challenges of Generative AI Tools for People with Disabilities} {Forgotten again: Addressing accessibility challenges of generative ai tools for people with disabilities}.{\BBCQ}
\newblock
\BIn{} \APACrefbtitle {Adjunct Proceedings of the 2024 Nordic Conference on Human-Computer Interaction.} {Adjunct proceedings of the 2024 nordic conference on human-computer interaction.}
\newblock
\APACaddressPublisher{New York, NY, USA}{Association for Computing Machinery}.
\newblock
\begin{APACrefDOI} \doi{10.1145/3677045.3685493} \end{APACrefDOI}
\PrintBackRefs{\CurrentBib}

\bibitem [\protect \citeauthoryear {%
Azaiz%
, Kiesler%
\BCBL {}\ \BBA {} Strickroth%
}{%
Azaiz%
\ \protect \BOthers {.}}{%
{\protect \APACyear {2024}}%
}]{%
azaiz2024feedbackgeneration}
\APACinsertmetastar {%
azaiz2024feedbackgeneration}%
\begin{APACrefauthors}%
Azaiz, I.%
, Kiesler, N.%
\BCBL {}\ \BBA {} Strickroth, S.%
\end{APACrefauthors}%
\unskip\
\newblock
\APACrefYearMonthDay{2024}{}{}.
\newblock
\APACrefbtitle {Feedback-Generation for Programming Exercises With GPT-4.} {Feedback-generation for programming exercises with gpt-4.}
\newblock
\APACaddressPublisher{New York, NY, USA}{Association for Computing Machinery}.
\newblock
\begin{APACrefDOI} \doi{10.1145/3649217.3653594} \end{APACrefDOI}
\PrintBackRefs{\CurrentBib}

\bibitem [\protect \citeauthoryear {%
Azaiz%
, Kiesler%
, Strickroth%
\BCBL {}\ \BBA {} Zhang%
}{%
Azaiz%
\ \protect \BOthers {.}}{%
{\protect \APACyear {2025}}%
}]{%
azaiz2025opensmallrigmarole}
\APACinsertmetastar {%
azaiz2025opensmallrigmarole}%
\begin{APACrefauthors}%
Azaiz, I.%
, Kiesler, N.%
, Strickroth, S.%
\BCBL {}\ \BBA {} Zhang, A.%
\end{APACrefauthors}%
\unskip\
\newblock
\APACrefYearMonthDay{2025}{}{}.
\newblock
\APACrefbtitle {Open, Small, Rigmarole -- Evaluating Llama 3.2 3B's Feedback for Programming Exercises.} {Open, small, rigmarole -- evaluating llama 3.2 3b's feedback for programming exercises.}
\newblock
\APACrefnote{accepted to the International Journal of Engineering Pedagogy (iJEP; eISSN: 2192-4880)}
\newblock
\begin{APACrefDOI} \doi{10.48550/arXiv.2504.01054} \end{APACrefDOI}
\PrintBackRefs{\CurrentBib}

\bibitem [\protect \citeauthoryear {%
Bengtsson%
\ \BBA {} Kaliff%
}{%
Bengtsson%
\ \BBA {} Kaliff%
}{%
{\protect \APACyear {2023}}%
}]{%
Bengtsson_Kaliff_2023}
\APACinsertmetastar {%
Bengtsson_Kaliff_2023}%
\begin{APACrefauthors}%
Bengtsson, D.%
\BCBT {}\ \BBA {} Kaliff, A.%
\end{APACrefauthors}%
\unskip\
\newblock
\APACrefYearMonthDay{2023}{}{}.
\newblock
\APACrefbtitle {Assessment Accuracy of a Large Language Model on Programming Assignments.} {Assessment accuracy of a large language model on programming assignments.}
\newblock
\begin{APACrefURL} \url{https://urn.kb.se/resolve?urn=urn:nbn:se:kth:diva-331000} \end{APACrefURL}
\PrintBackRefs{\CurrentBib}

\bibitem [\protect \citeauthoryear {%
Bhowmick%
\ \BBA {} Li%
}{%
Bhowmick%
\ \BBA {} Li%
}{%
{\protect \APACyear {2025}}%
}]{%
bhowmick2025experience}
\APACinsertmetastar {%
bhowmick2025experience}%
\begin{APACrefauthors}%
Bhowmick, S.%
\BCBT {}\ \BBA {} Li, H.%
\end{APACrefauthors}%
\unskip\
\newblock
\APACrefYearMonthDay{2025}{}{}.
\newblock
{\BBOQ}\APACrefatitle {Experience Report on Using LANTERN in Teaching Relational Query Processing} {Experience report on using lantern in teaching relational query processing}.{\BBCQ}
\newblock
\BIn{} \APACrefbtitle {Proceedings of the 56th ACM Technical Symposium on Computer Science Education V. 1} {Proceedings of the 56th acm technical symposium on computer science education v. 1}\ (\BPG~123–129).
\newblock
\APACaddressPublisher{New York, NY, USA}{Association for Computing Machinery}.
\newblock
\begin{APACrefDOI} \doi{10.1145/3641554.3701812} \end{APACrefDOI}
\PrintBackRefs{\CurrentBib}

\bibitem [\protect \citeauthoryear {%
Boddy%
}{%
Boddy%
}{%
{\protect \APACyear {2016}}%
}]{%
boddy2016sample}
\APACinsertmetastar {%
boddy2016sample}%
\begin{APACrefauthors}%
Boddy, C\BPBI R.%
\end{APACrefauthors}%
\unskip\
\newblock
\APACrefYearMonthDay{2016}{}{}.
\newblock
{\BBOQ}\APACrefatitle {Sample size for qualitative research} {Sample size for qualitative research}.{\BBCQ}
\newblock
\APACjournalVolNumPages{Qualitative market research: An international journal}{19}{4}{426--432}.
\PrintBackRefs{\CurrentBib}

\bibitem [\protect \citeauthoryear {%
Du~Boulay%
}{%
Du~Boulay%
}{%
{\protect \APACyear {1986}}%
}]{%
duboulay1986some}
\APACinsertmetastar {%
duboulay1986some}%
\begin{APACrefauthors}%
Du~Boulay, B.%
\end{APACrefauthors}%
\unskip\
\newblock
\APACrefYearMonthDay{1986}{}{}.
\newblock
{\BBOQ}\APACrefatitle {{Some difficulties of learning to program}} {{Some difficulties of learning to program}}.{\BBCQ}
\newblock
\APACjournalVolNumPages{{Journal of Educational Computing Research}}{2}{1}{57--73}.
\newblock
\begin{APACrefDOI} \doi{10.2190/3LFX-9RRF-67T8-UVK9} \end{APACrefDOI}
\PrintBackRefs{\CurrentBib}

\bibitem [\protect \citeauthoryear {%
Ebert%
\ \BBA {} Ring%
}{%
Ebert%
\ \BBA {} Ring%
}{%
{\protect \APACyear {2016}}%
}]{%
Ebert2016}
\APACinsertmetastar {%
Ebert2016}%
\begin{APACrefauthors}%
Ebert, M.%
\BCBT {}\ \BBA {} Ring, M.%
\end{APACrefauthors}%
\unskip\
\newblock
\APACrefYearMonthDay{2016}{}{}.
\newblock
{\BBOQ}\APACrefatitle {A presentation framework for programming in programing lectures} {A presentation framework for programming in programing lectures}.{\BBCQ}
\newblock
\BIn{} \APACrefbtitle {Proc. EDUCON} {Proc. educon}\ (\BPGS\ 369--374).
\PrintBackRefs{\CurrentBib}

\bibitem [\protect \citeauthoryear {%
Forden%
, Schneider%
, Gebhard%
, Islam~Molla%
\BCBL {}\ \BBA {} Brylow%
}{%
Forden%
\ \protect \BOthers {.}}{%
{\protect \APACyear {2025}}%
}]{%
forden2025unlocking}
\APACinsertmetastar {%
forden2025unlocking}%
\begin{APACrefauthors}%
Forden, J.%
, Schneider, M.%
, Gebhard, A.%
, Islam~Molla, M\BPBI T.%
\BCBL {}\ \BBA {} Brylow, D.%
\end{APACrefauthors}%
\unskip\
\newblock
\APACrefYearMonthDay{2025}{}{}.
\newblock
{\BBOQ}\APACrefatitle {Unlocking Student Potential With TA-Bot: Timely Submissions and Improved Code Style} {Unlocking student potential with ta-bot: Timely submissions and improved code style}.{\BBCQ}
\newblock
\BIn{} \APACrefbtitle {Proceedings of the 56th ACM Technical Symposium on Computer Science Education V. 1} {Proceedings of the 56th acm technical symposium on computer science education v. 1}\ (\BPG~346–352).
\newblock
\APACaddressPublisher{New York, NY, USA}{Association for Computing Machinery}.
\newblock
\begin{APACrefDOI} \doi{10.1145/3641554.3701955} \end{APACrefDOI}
\PrintBackRefs{\CurrentBib}

\bibitem [\protect \citeauthoryear {%
Geng%
, Zhang%
, Pientka%
\BCBL {}\ \BBA {} Si%
}{%
Geng%
\ \protect \BOthers {.}}{%
{\protect \APACyear {2023}}%
}]{%
geng2023chatgpt}
\APACinsertmetastar {%
geng2023chatgpt}%
\begin{APACrefauthors}%
Geng, C.%
, Zhang, Y.%
, Pientka, B.%
\BCBL {}\ \BBA {} Si, X.%
\end{APACrefauthors}%
\unskip\
\newblock
\APACrefYearMonthDay{2023}{}{}.
\newblock
\APACrefbtitle {{Can ChatGPT Pass An Introductory Level Functional Language Programming Course?}} {{Can ChatGPT Pass An Introductory Level Functional Language Programming Course?}}
\PrintBackRefs{\CurrentBib}

\bibitem [\protect \citeauthoryear {%
Gill%
\ \protect \BOthers {.}}{%
Gill%
\ \protect \BOthers {.}}{%
{\protect \APACyear {2024}}%
}]{%
gill2024transformative}
\APACinsertmetastar {%
gill2024transformative}%
\begin{APACrefauthors}%
Gill, S\BPBI S.%
, Xu, M.%
, Patros, P.%
, Wu, H.%
, Kaur, R.%
, Kaur, K.%
\BDBL {}Buyya, R.%
\end{APACrefauthors}%
\unskip\
\newblock
\APACrefYearMonthDay{2024}{}{}.
\newblock
{\BBOQ}\APACrefatitle {Transformative effects of ChatGPT on modern education: Emerging Era of AI Chatbots} {Transformative effects of chatgpt on modern education: Emerging era of ai chatbots}.{\BBCQ}
\newblock
\APACjournalVolNumPages{Internet of Things and Cyber-Physical Systems}{4}{}{19-23}.
\newblock
\begin{APACrefDOI} \doi{10.1016/j.iotcps.2023.06.002} \end{APACrefDOI}
\PrintBackRefs{\CurrentBib}

\bibitem [\protect \citeauthoryear {%
Jacobs%
\ \BBA {} Jaschke%
}{%
Jacobs%
\ \BBA {} Jaschke%
}{%
{\protect \APACyear {2024}}%
}]{%
jacobs2024evaluating}
\APACinsertmetastar {%
jacobs2024evaluating}%
\begin{APACrefauthors}%
Jacobs, S.%
\BCBT {}\ \BBA {} Jaschke, S.%
\end{APACrefauthors}%
\unskip\
\newblock
\APACrefYearMonthDay{2024}{{\APACmonth{05}}}{}.
\newblock
{\BBOQ}\APACrefatitle {Evaluating the {{Application}} of {{Large Language Models}} to {{Generate Feedback}} in {{Programming Education}}} {Evaluating the {{Application}} of {{Large Language Models}} to {{Generate Feedback}} in {{Programming Education}}}.{\BBCQ}
\newblock
\BIn{} \APACrefbtitle {2024 {{IEEE Global Engineering Education Conference}} ({{EDUCON}})} {2024 {{IEEE Global Engineering Education Conference}} ({{EDUCON}})}\ (\BPGS\ 1--5).
\newblock
\APACaddressPublisher{New York}{IEEE}.
\newblock
\begin{APACrefDOI} \doi{10.1109/EDUCON60312.2024.10578838} \end{APACrefDOI}
\PrintBackRefs{\CurrentBib}

\bibitem [\protect \citeauthoryear {%
Jacobs%
, Kempf%
\BCBL {}\ \BBA {} Kiesler%
}{%
Jacobs%
, Kempf%
\BCBL {}\ \BBA {} Kiesler%
}{%
{\protect \APACyear {2025}}%
}]{%
jacobs2025thatsnotthefeedback}
\APACinsertmetastar {%
jacobs2025thatsnotthefeedback}%
\begin{APACrefauthors}%
Jacobs, S.%
, Kempf, M.%
\BCBL {}\ \BBA {} Kiesler, N.%
\end{APACrefauthors}%
\unskip\
\newblock
\APACrefYearMonthDay{2025}{}{}.
\newblock
{\BBOQ}\APACrefatitle {That's Not the Feedback I Need! -- Student Engagement with GenAI Feedback in the Tutor Kai} {That's not the feedback i need! -- student engagement with genai feedback in the tutor kai}.{\BBCQ}.
\newblock
\begin{APACrefURL} \url{https://arxiv.org/abs/2506.20433} \end{APACrefURL}
\PrintBackRefs{\CurrentBib}

\bibitem [\protect \citeauthoryear {%
Jacobs%
, Peters%
, Jaschke%
\BCBL {}\ \BBA {} Kiesler%
}{%
Jacobs%
, Peters%
\BCBL {}\ \protect \BOthers {.}}{%
{\protect \APACyear {2025}}%
}]{%
jacobs2025unlimited}
\APACinsertmetastar {%
jacobs2025unlimited}%
\begin{APACrefauthors}%
Jacobs, S.%
, Peters, H.%
, Jaschke, S.%
\BCBL {}\ \BBA {} Kiesler, N.%
\end{APACrefauthors}%
\unskip\
\newblock
\APACrefYearMonthDay{2025}{}{}.
\newblock
\APACrefbtitle {Unlimited Practice Opportunities: Automated Generation of Comprehensive, Personalized Programming Tasks.} {Unlimited practice opportunities: Automated generation of comprehensive, personalized programming tasks.}
\newblock
\APACrefnote{accepted at ITiCSE 2025, https://doi.org/10.1145/3724363.3729089}
\newblock
\begin{APACrefDOI} \doi{10.48550/arXiv.2503.11704} \end{APACrefDOI}
\PrintBackRefs{\CurrentBib}

\bibitem [\protect \citeauthoryear {%
Kazemitabaar%
\ \protect \BOthers {.}}{%
Kazemitabaar%
\ \protect \BOthers {.}}{%
{\protect \APACyear {2024}}%
}]{%
kazemitabaar2024codeaid}
\APACinsertmetastar {%
kazemitabaar2024codeaid}%
\begin{APACrefauthors}%
Kazemitabaar, M.%
, Ye, R.%
, Wang, X.%
, Henley, A\BPBI Z.%
, Denny, P.%
, Craig, M.%
\BCBL {}\ \BBA {} Grossman, T.%
\end{APACrefauthors}%
\unskip\
\newblock
\APACrefYearMonthDay{2024}{}{}.
\newblock
{\BBOQ}\APACrefatitle {CodeAid: Evaluating a Classroom Deployment of an LLM-based Programming Assistant that Balances Student and Educator Needs} {Codeaid: Evaluating a classroom deployment of an llm-based programming assistant that balances student and educator needs}.{\BBCQ}
\newblock
\BIn{} \APACrefbtitle {Proceedings of the CHI Conference on Human Factors in Computing Systems.} {Proceedings of the chi conference on human factors in computing systems.}
\newblock
\APACaddressPublisher{New York, USA}{ACM}.
\newblock
\begin{APACrefDOI} \doi{10.1145/3613904.3642773} \end{APACrefDOI}
\PrintBackRefs{\CurrentBib}

\bibitem [\protect \citeauthoryear {%
Keller%
}{%
Keller%
}{%
{\protect \APACyear {1968}}%
}]{%
keller1968good}
\APACinsertmetastar {%
keller1968good}%
\begin{APACrefauthors}%
Keller, F\BPBI S.%
\end{APACrefauthors}%
\unskip\
\newblock
\APACrefYearMonthDay{1968}{}{}.
\newblock
{\BBOQ}\APACrefatitle {Good-bye, teacher...} {Good-bye, teacher...}{\BBCQ}
\newblock
\APACjournalVolNumPages{Journal of applied behavior analysis}{1}{1}{79}.
\PrintBackRefs{\CurrentBib}

\bibitem [\protect \citeauthoryear {%
Keuning%
, Jeuring%
\BCBL {}\ \BBA {} Heeren%
}{%
Keuning%
\ \protect \BOthers {.}}{%
{\protect \APACyear {2018}}%
}]{%
keuning2018}
\APACinsertmetastar {%
keuning2018}%
\begin{APACrefauthors}%
Keuning, H.%
, Jeuring, J.%
\BCBL {}\ \BBA {} Heeren, B.%
\end{APACrefauthors}%
\unskip\
\newblock
\APACrefYearMonthDay{2018}{9}{}.
\newblock
{\BBOQ}\APACrefatitle {A Systematic Literature Review of Automated Feedback Generation for Programming Exercises} {A systematic literature review of automated feedback generation for programming exercises}.{\BBCQ}
\newblock
\APACjournalVolNumPages{ACM Trans. Comput. Educ.}{19}{1}{}.
\newblock
\begin{APACrefDOI} \doi{10.1145/3231711} \end{APACrefDOI}
\PrintBackRefs{\CurrentBib}

\bibitem [\protect \citeauthoryear {%
Kiesler%
}{%
Kiesler%
}{%
{\protect \APACyear {2022}}%
}]{%
kiesler_diss_2022}
\APACinsertmetastar {%
kiesler_diss_2022}%
\begin{APACrefauthors}%
Kiesler, N.%
\end{APACrefauthors}%
\unskip\
\newblock
\APACrefYear{2022}.
\unskip\
\newblock
\APACrefbtitle {Kompetenzförderung in der Programmierausbildung durch Modellierung von Kompetenzen und informativem Feedback} {Kompetenzförderung in der programmierausbildung durch modellierung von kompetenzen und informativem feedback}\ \APACtypeAddressSchool {Dissertation}{}{}.
\unskip\
\newblock
\APACaddressSchool {Frankfurt am Main}{Johann Wolfgang Goethe-Universität}.
\unskip\
\newblock
\APACrefnote{Fachbereich Informatik und Mathematik}
\PrintBackRefs{\CurrentBib}

\bibitem [\protect \citeauthoryear {%
Kiesler%
}{%
Kiesler%
}{%
{\protect \APACyear {2024}}%
}]{%
kiesler2024modeling}
\APACinsertmetastar {%
kiesler2024modeling}%
\begin{APACrefauthors}%
Kiesler, N.%
\end{APACrefauthors}%
\unskip\
\newblock
\APACrefYear{2024}.
\newblock
\APACrefbtitle {Modeling Programming Competency: A Qualitative Analysis} {Modeling programming competency: A qualitative analysis}.
\newblock
\APACaddressPublisher{Cham}{Springer International Publishing}.
\newblock
\begin{APACrefDOI} \doi{10.1007/978-3-031-47148-3} \end{APACrefDOI}
\PrintBackRefs{\CurrentBib}

\bibitem [\protect \citeauthoryear {%
Kiesler%
, Lohr%
\BCBL {}\ \BBA {} Keuning%
}{%
Kiesler%
\ \protect \BOthers {.}}{%
{\protect \APACyear {2024}}%
}]{%
kiesler2023exploring}
\APACinsertmetastar {%
kiesler2023exploring}%
\begin{APACrefauthors}%
Kiesler, N.%
, Lohr, D.%
\BCBL {}\ \BBA {} Keuning, H.%
\end{APACrefauthors}%
\unskip\
\newblock
\APACrefYearMonthDay{2024}{}{}.
\newblock
{\BBOQ}\APACrefatitle {Exploring the Potential of Large Language Models to Generate Formative Programming Feedback} {Exploring the potential of large language models to generate formative programming feedback}.{\BBCQ}
\newblock
\BIn{} \APACrefbtitle {2023 IEEE Frontiers in Education Conference (FIE)} {2023 ieee frontiers in education conference (fie)}\ (\BPG~1-5).
\newblock
\begin{APACrefDOI} \doi{10.1109/FIE58773.2023.10343457} \end{APACrefDOI}
\PrintBackRefs{\CurrentBib}

\bibitem [\protect \citeauthoryear {%
Kiesler%
\ \BBA {} Schiffner%
}{%
Kiesler%
\ \BBA {} Schiffner%
}{%
{\protect \APACyear {2023}}%
{\protect \APACexlab {{\protect \BCnt {1}}}}}]{%
kiesler2023large}
\APACinsertmetastar {%
kiesler2023large}%
\begin{APACrefauthors}%
Kiesler, N.%
\BCBT {}\ \BBA {} Schiffner, D.%
\end{APACrefauthors}%
\unskip\
\newblock
\APACrefYearMonthDay{2023{\protect \BCnt {1}}}{}{}.
\newblock
\APACrefbtitle {Large Language Models in Introductory Programming Education: ChatGPT's Performance and Implications for Assessments.} {Large language models in introductory programming education: Chatgpt's performance and implications for assessments.}
\newblock
\begin{APACrefDOI} \doi{10.48550/arXiv.2308.08572} \end{APACrefDOI}
\PrintBackRefs{\CurrentBib}

\bibitem [\protect \citeauthoryear {%
Kiesler%
\ \BBA {} Schiffner%
}{%
Kiesler%
\ \BBA {} Schiffner%
}{%
{\protect \APACyear {2023}}%
{\protect \APACexlab {{\protect \BCnt {2}}}}}]{%
kiesler2023why}
\APACinsertmetastar {%
kiesler2023why}%
\begin{APACrefauthors}%
Kiesler, N.%
\BCBT {}\ \BBA {} Schiffner, D.%
\end{APACrefauthors}%
\unskip\
\newblock
\APACrefYearMonthDay{2023{\protect \BCnt {2}}}{}{}.
\newblock
{\BBOQ}\APACrefatitle {{Why We Need Open Data in Computer Science Education Research}} {{Why We Need Open Data in Computer Science Education Research}}.{\BBCQ}
\newblock
\BIn{} \APACrefbtitle {Proceedings of the 2023 Conference on Innovation and Technology in Computer Science Education V. 1} {Proceedings of the 2023 conference on innovation and technology in computer science education v. 1}\ (\BPG~348–353).
\newblock
\APACaddressPublisher{New York}{ACM}.
\newblock
\begin{APACrefDOI} \doi{10.1145/3587102.3588860} \end{APACrefDOI}
\PrintBackRefs{\CurrentBib}

\bibitem [\protect \citeauthoryear {%
Kiesler%
\ \protect \BOthers {.}}{%
Kiesler%
\ \protect \BOthers {.}}{%
{\protect \APACyear {2025}}%
}]{%
kiesler2025role}
\APACinsertmetastar {%
kiesler2025role}%
\begin{APACrefauthors}%
Kiesler, N.%
, Smith, J.%
, Leinonen, J.%
, Fox, A.%
, MacNeil, S.%
\BCBL {}\ \BBA {} Ihantola, P.%
\end{APACrefauthors}%
\unskip\
\newblock
\APACrefYearMonthDay{2025}{}{}.
\newblock
\APACrefbtitle {The Role of Generative AI in Software Student CollaborAItion.} {The role of generative ai in software student collaboraition.}
\newblock
\APACaddressPublisher{New York, NY, USA}{Association for Computing Machinery}.
\newblock
\begin{APACrefDOI} \doi{10.1145/3724363.3729040} \end{APACrefDOI}
\PrintBackRefs{\CurrentBib}

\bibitem [\protect \citeauthoryear {%
Leinonen%
\ \protect \BOthers {.}}{%
Leinonen%
\ \protect \BOthers {.}}{%
{\protect \APACyear {2023}}%
}]{%
Leinonen2023}
\APACinsertmetastar {%
Leinonen2023}%
\begin{APACrefauthors}%
Leinonen, J.%
, Hellas, A.%
, Sarsa, S.%
, Reeves, B.%
, Denny, P.%
, Prather, J.%
\BCBL {}\ \BBA {} Becker, B\BPBI A.%
\end{APACrefauthors}%
\unskip\
\newblock
\APACrefYearMonthDay{2023}{{\APACmonth{03}}}{}.
\newblock
{\BBOQ}\APACrefatitle {Using Large Language Models to Enhance Programming Error Messages} {Using large language models to enhance programming error messages}.{\BBCQ}
\newblock
\BIn{} \APACrefbtitle {Proc. SIGCSE.} {Proc. sigcse.}
\newblock
\APACaddressPublisher{}{ACM}.
\newblock
\begin{APACrefDOI} \doi{10.1145/3545945.3569770} \end{APACrefDOI}
\PrintBackRefs{\CurrentBib}

\bibitem [\protect \citeauthoryear {%
Liffiton%
, Sheese%
, Savelka%
\BCBL {}\ \BBA {} Denny%
}{%
Liffiton%
\ \protect \BOthers {.}}{%
{\protect \APACyear {2024}}%
}]{%
liffiton2023codehelp}
\APACinsertmetastar {%
liffiton2023codehelp}%
\begin{APACrefauthors}%
Liffiton, M.%
, Sheese, B\BPBI E.%
, Savelka, J.%
\BCBL {}\ \BBA {} Denny, P.%
\end{APACrefauthors}%
\unskip\
\newblock
\APACrefYearMonthDay{2024}{}{}.
\newblock
{\BBOQ}\APACrefatitle {CodeHelp: Using Large Language Models with Guardrails for Scalable Support in Programming Classes} {Codehelp: Using large language models with guardrails for scalable support in programming classes}.{\BBCQ}
\newblock
\BIn{} \APACrefbtitle {Proceedings of the 23rd Koli Calling International Conference on Computing Education Research.} {Proceedings of the 23rd koli calling international conference on computing education research.}
\newblock
\APACaddressPublisher{New York}{ACM}.
\newblock
\begin{APACrefDOI} \doi{10.1145/3631802.3631830} \end{APACrefDOI}
\PrintBackRefs{\CurrentBib}

\bibitem [\protect \citeauthoryear {%
Liu%
\ \protect \BOthers {.}}{%
Liu%
\ \protect \BOthers {.}}{%
{\protect \APACyear {2024}}%
}]{%
liu2024teaching}
\APACinsertmetastar {%
liu2024teaching}%
\begin{APACrefauthors}%
Liu, R.%
, Zenke, C.%
, Liu, C.%
, Holmes, A.%
, Thornton, P.%
\BCBL {}\ \BBA {} Malan, D\BPBI J.%
\end{APACrefauthors}%
\unskip\
\newblock
\APACrefYearMonthDay{2024}{}{}.
\newblock
{\BBOQ}\APACrefatitle {Teaching CS50 with AI: Leveraging Generative Artificial Intelligence in Computer Science Education} {Teaching cs50 with ai: Leveraging generative artificial intelligence in computer science education}.{\BBCQ}
\newblock
\BIn{} \APACrefbtitle {Proceedings of the 55th ACM Technical Symposium on Computer Science Education V. 1} {Proceedings of the 55th acm technical symposium on computer science education v. 1}\ (\BPG~750–756).
\newblock
\begin{APACrefDOI} \doi{10.1145/3626252.3630938} \end{APACrefDOI}
\PrintBackRefs{\CurrentBib}

\bibitem [\protect \citeauthoryear {%
Liu%
\ \protect \BOthers {.}}{%
Liu%
\ \protect \BOthers {.}}{%
{\protect \APACyear {2025}}%
}]{%
liu2025improving}
\APACinsertmetastar {%
liu2025improving}%
\begin{APACrefauthors}%
Liu, R.%
, Zhao, J.%
, Xu, B.%
, Perez, C.%
, Zhukovets, Y.%
\BCBL {}\ \BBA {} Malan, D\BPBI J.%
\end{APACrefauthors}%
\unskip\
\newblock
\APACrefYearMonthDay{2025}{}{}.
\newblock
{\BBOQ}\APACrefatitle {Improving AI in CS50: Leveraging Human Feedback for Better Learning} {Improving ai in cs50: Leveraging human feedback for better learning}.{\BBCQ}
\newblock
\BIn{} \APACrefbtitle {Proceedings of the 56th ACM Technical Symposium on Computer Science Education V. 1} {Proceedings of the 56th acm technical symposium on computer science education v. 1}\ (\BPG~715–721).
\newblock
\APACaddressPublisher{New York, NY, USA}{ACM}.
\newblock
\begin{APACrefDOI} \doi{10.1145/3641554.3701945} \end{APACrefDOI}
\PrintBackRefs{\CurrentBib}

\bibitem [\protect \citeauthoryear {%
Lohr%
, Keuning%
\BCBL {}\ \BBA {} Kiesler%
}{%
Lohr%
\ \protect \BOthers {.}}{%
{\protect \APACyear {2025}}%
}]{%
lohr2025youre}
\APACinsertmetastar {%
lohr2025youre}%
\begin{APACrefauthors}%
Lohr, D.%
, Keuning, H.%
\BCBL {}\ \BBA {} Kiesler, N.%
\end{APACrefauthors}%
\unskip\
\newblock
\APACrefYearMonthDay{2025}{}{}.
\newblock
{\BBOQ}\APACrefatitle {You're ({{Not}}) {{My Type}} -- {{Can LLMs Generate Feedback}} of {{Specific Types}} for {{Introductory Programming Tasks}}?} {You're ({{Not}}) {{My Type}} -- {{Can LLMs Generate Feedback}} of {{Specific Types}} for {{Introductory Programming Tasks}}?}{\BBCQ}
\newblock
\APACjournalVolNumPages{Journal of Computer Assisted Learning}{}{}{}.
\newblock
\begin{APACrefDOI} \doi{10.1111/jcal.13107} \end{APACrefDOI}
\PrintBackRefs{\CurrentBib}

\bibitem [\protect \citeauthoryear {%
Luxton-Reilly%
}{%
Luxton-Reilly%
}{%
{\protect \APACyear {2016}}%
}]{%
luxton-reilly2016}
\APACinsertmetastar {%
luxton-reilly2016}%
\begin{APACrefauthors}%
Luxton-Reilly, A.%
\end{APACrefauthors}%
\unskip\
\newblock
\APACrefYearMonthDay{2016}{}{}.
\newblock
{\BBOQ}\APACrefatitle {{Learning to Program is Easy}} {{Learning to Program is Easy}}.{\BBCQ}
\newblock
\BIn{} \APACrefbtitle {{Proc. ITiCSE}} {{Proc. ITiCSE}}\ (\BPGS\ 284--289).
\newblock
\begin{APACrefDOI} \doi{10.1145/2899415.2899432} \end{APACrefDOI}
\PrintBackRefs{\CurrentBib}

\bibitem [\protect \citeauthoryear {%
Luxton-Reilly%
\ \protect \BOthers {.}}{%
Luxton-Reilly%
\ \protect \BOthers {.}}{%
{\protect \APACyear {2018}}%
}]{%
Luxton-Reilly2018}
\APACinsertmetastar {%
Luxton-Reilly2018}%
\begin{APACrefauthors}%
Luxton-Reilly, A.%
, Simon%
, Albluwi, I.%
, Becker, B\BPBI A.%
, Giannakos, M.%
, Kumar, A\BPBI N.%
\BDBL {}Szabo, C.%
\end{APACrefauthors}%
\unskip\
\newblock
\APACrefYearMonthDay{2018}{}{}.
\newblock
{\BBOQ}\APACrefatitle {{Introductory Programming: A Systematic Literature Review}} {{Introductory Programming: A Systematic Literature Review}}.{\BBCQ}
\newblock
\BIn{} \APACrefbtitle {{Proc. ITiCSE}} {{Proc. ITiCSE}}\ (\BPGS\ 55--106).
\newblock
\APACaddressPublisher{New York}{ACM}.
\newblock
\begin{APACrefDOI} \doi{10.1145/3293881.3295779} \end{APACrefDOI}
\PrintBackRefs{\CurrentBib}

\bibitem [\protect \citeauthoryear {%
Lyu%
, Wang%
, Chung%
, Sun%
\BCBL {}\ \BBA {} Zhang%
}{%
Lyu%
\ \protect \BOthers {.}}{%
{\protect \APACyear {2024}}%
}]{%
lyu2024evaluating}
\APACinsertmetastar {%
lyu2024evaluating}%
\begin{APACrefauthors}%
Lyu, W.%
, Wang, Y.%
, Chung, T\BPBI R.%
, Sun, Y.%
\BCBL {}\ \BBA {} Zhang, Y.%
\end{APACrefauthors}%
\unskip\
\newblock
\APACrefYearMonthDay{2024}{}{}.
\newblock
{\BBOQ}\APACrefatitle {Evaluating the Effectiveness of LLMs in Introductory Computer Science Education: A Semester-Long Field Study} {Evaluating the effectiveness of llms in introductory computer science education: A semester-long field study}.{\BBCQ}
\newblock
\APACjournalVolNumPages{arXiv:2404.13414}{}{}{}.
\PrintBackRefs{\CurrentBib}

\bibitem [\protect \citeauthoryear {%
MacNeil%
\ \protect \BOthers {.}}{%
MacNeil%
\ \protect \BOthers {.}}{%
{\protect \APACyear {2023}}%
}]{%
macneil2022experiences}
\APACinsertmetastar {%
macneil2022experiences}%
\begin{APACrefauthors}%
MacNeil, S.%
, Tran, A.%
, Hellas, A.%
, Kim, J.%
, Sarsa, S.%
, Denny, P.%
\BDBL {}Leinonen, J.%
\end{APACrefauthors}%
\unskip\
\newblock
\APACrefYearMonthDay{2023}{}{}.
\newblock
{\BBOQ}\APACrefatitle {{Experiences from Using Code Explanations Generated by Large Language Models in a Web Software Development E-Book}} {{Experiences from Using Code Explanations Generated by Large Language Models in a Web Software Development E-Book}}.{\BBCQ}
\newblock
\BIn{} \APACrefbtitle {Proc. SIGCSE TS} {Proc. sigcse ts}\ (\BPG~931–937).
\newblock
\begin{APACrefDOI} \doi{10.1145/3545945.3569785} \end{APACrefDOI}
\PrintBackRefs{\CurrentBib}

\bibitem [\protect \citeauthoryear {%
Narciss%
}{%
Narciss%
}{%
{\protect \APACyear {2006}}%
}]{%
narciss2006}
\APACinsertmetastar {%
narciss2006}%
\begin{APACrefauthors}%
Narciss, S.%
\end{APACrefauthors}%
\unskip\
\newblock
\APACrefYear{2006}.
\newblock
\APACrefbtitle {Informatives Tutorielles Feedback: Entwicklungs- und Evaluationsprinzipien auf der Basis instruktionspsychologischer Erkenntnisse} {Informatives tutorielles feedback: Entwicklungs- und evaluationsprinzipien auf der basis instruktionspsychologischer erkenntnisse}.
\newblock
\APACaddressPublisher{M\"{u}nster}{Waxmann Verlag}.
\PrintBackRefs{\CurrentBib}

\bibitem [\protect \citeauthoryear {%
Narciss%
}{%
Narciss%
}{%
{\protect \APACyear {2008}}%
}]{%
narciss2008feedback}
\APACinsertmetastar {%
narciss2008feedback}%
\begin{APACrefauthors}%
Narciss, S.%
\end{APACrefauthors}%
\unskip\
\newblock
\APACrefYearMonthDay{2008}{}{}.
\newblock
{\BBOQ}\APACrefatitle {Feedback strategies for interactive learning tasks} {Feedback strategies for interactive learning tasks}.{\BBCQ}
\newblock
\APACjournalVolNumPages{Handbook of research on educational communications and technology}{3}{}{125--144}.
\PrintBackRefs{\CurrentBib}

\bibitem [\protect \citeauthoryear {%
Nelson%
, Doup\'{e}%
\BCBL {}\ \BBA {} Shoshitaishvili%
}{%
Nelson%
\ \protect \BOthers {.}}{%
{\protect \APACyear {2025}}%
}]{%
nelson2025sensai}
\APACinsertmetastar {%
nelson2025sensai}%
\begin{APACrefauthors}%
Nelson, C.%
, Doup\'{e}, A.%
\BCBL {}\ \BBA {} Shoshitaishvili, Y.%
\end{APACrefauthors}%
\unskip\
\newblock
\APACrefYearMonthDay{2025}{}{}.
\newblock
{\BBOQ}\APACrefatitle {SENSAI: Large Language Models as Applied Cybersecurity Tutors} {Sensai: Large language models as applied cybersecurity tutors}.{\BBCQ}
\newblock
\BIn{} \APACrefbtitle {Proceedings of the 56th ACM Technical Symposium on Computer Science Education V. 1} {Proceedings of the 56th acm technical symposium on computer science education v. 1}\ (\BPG~833–839).
\newblock
\APACaddressPublisher{New York, NY, USA}{Association for Computing Machinery}.
\newblock
\begin{APACrefDOI} \doi{10.1145/3641554.3701801} \end{APACrefDOI}
\PrintBackRefs{\CurrentBib}

\bibitem [\protect \citeauthoryear {%
Petersen%
, Craig%
, Campbell%
\BCBL {}\ \BBA {} Tafliovich%
}{%
Petersen%
\ \protect \BOthers {.}}{%
{\protect \APACyear {2016}}%
}]{%
petersen2016revisiting}
\APACinsertmetastar {%
petersen2016revisiting}%
\begin{APACrefauthors}%
Petersen, A.%
, Craig, M.%
, Campbell, J.%
\BCBL {}\ \BBA {} Tafliovich, A.%
\end{APACrefauthors}%
\unskip\
\newblock
\APACrefYearMonthDay{2016}{}{}.
\newblock
{\BBOQ}\APACrefatitle {Revisiting why students drop CS1} {Revisiting why students drop cs1}.{\BBCQ}
\newblock
\BIn{} \APACrefbtitle {Proc. Koli Calling} {Proc. koli calling}\ (\BPGS\ 71--80).
\newblock
\begin{APACrefDOI} \doi{10.1145/2999541.2999552} \end{APACrefDOI}
\PrintBackRefs{\CurrentBib}

\bibitem [\protect \citeauthoryear {%
Phung%
\ \protect \BOthers {.}}{%
Phung%
\ \protect \BOthers {.}}{%
{\protect \APACyear {2023}}%
}]{%
phung2023generating}
\APACinsertmetastar {%
phung2023generating}%
\begin{APACrefauthors}%
Phung, T.%
, Cambronero, J.%
, Gulwani, S.%
, Kohn, T.%
, Majumdar, R.%
, Singla, A.%
\BCBL {}\ \BBA {} Soares, G.%
\end{APACrefauthors}%
\unskip\
\newblock
\APACrefYearMonthDay{2023}{}{}.
\newblock
\APACrefbtitle {{Generating High-Precision Feedback for Programming Syntax Errors using Large Language Models}.} {{Generating High-Precision Feedback for Programming Syntax Errors using Large Language Models}.}
\PrintBackRefs{\CurrentBib}

\bibitem [\protect \citeauthoryear {%
Prather%
\ \protect \BOthers {.}}{%
Prather%
\ \protect \BOthers {.}}{%
{\protect \APACyear {2023}}%
}]{%
prather2023wgfullreport}
\APACinsertmetastar {%
prather2023wgfullreport}%
\begin{APACrefauthors}%
Prather, J.%
, Denny, P.%
, Leinonen, J.%
, Becker, B\BPBI A.%
, Albluwi, I.%
, Craig, M.%
\BDBL {}Savelka, J.%
\end{APACrefauthors}%
\unskip\
\newblock
\APACrefYearMonthDay{2023}{}{}.
\newblock
{\BBOQ}\APACrefatitle {The Robots Are Here: Navigating the Generative AI Revolution in Computing Education} {The robots are here: Navigating the generative ai revolution in computing education}.{\BBCQ}
\newblock
\BIn{} \APACrefbtitle {Proceedings of the 2023 Working Group Reports on Innovation and Technology in Computer Science Education} {Proceedings of the 2023 working group reports on innovation and technology in computer science education}\ (\BPG~108–159).
\newblock
\APACaddressPublisher{New York}{ACM}.
\newblock
\begin{APACrefDOI} \doi{10.1145/3623762.3633499} \end{APACrefDOI}
\PrintBackRefs{\CurrentBib}

\bibitem [\protect \citeauthoryear {%
Prather%
\ \protect \BOthers {.}}{%
Prather%
\ \protect \BOthers {.}}{%
{\protect \APACyear {2025}}%
}]{%
prather2025beyond}
\APACinsertmetastar {%
prather2025beyond}%
\begin{APACrefauthors}%
Prather, J.%
, Leinonen, J.%
, Kiesler, N.%
, Gorson~Benario, J.%
, Lau, S.%
, MacNeil, S.%
\BDBL {}Zingaro, D.%
\end{APACrefauthors}%
\unskip\
\newblock
\APACrefYearMonthDay{2025}{}{}.
\newblock
{\BBOQ}\APACrefatitle {Beyond the Hype: A Comprehensive Review of Current Trends in Generative AI Research, Teaching Practices, and Tools} {Beyond the hype: A comprehensive review of current trends in generative ai research, teaching practices, and tools}.{\BBCQ}
\newblock
\BIn{} \APACrefbtitle {2024 Working Group Reports on Innovation and Technology in Computer Science Education} {2024 working group reports on innovation and technology in computer science education}\ (\BPG~300–338).
\newblock
\APACaddressPublisher{New York, NY, USA}{Association for Computing Machinery}.
\newblock
\begin{APACrefDOI} \doi{10.1145/3689187.3709614} \end{APACrefDOI}
\PrintBackRefs{\CurrentBib}

\bibitem [\protect \citeauthoryear {%
Renzella%
, Vassar%
, Lee~Solano%
\BCBL {}\ \BBA {} Taylor%
}{%
Renzella%
\ \protect \BOthers {.}}{%
{\protect \APACyear {2025}}%
}]{%
renzella2025compiler}
\APACinsertmetastar {%
renzella2025compiler}%
\begin{APACrefauthors}%
Renzella, J.%
, Vassar, A.%
, Lee~Solano, L.%
\BCBL {}\ \BBA {} Taylor, A.%
\end{APACrefauthors}%
\unskip\
\newblock
\APACrefYearMonthDay{2025}{}{}.
\newblock
{\BBOQ}\APACrefatitle {Compiler-Integrated, Conversational AI for Debugging CS1 Programs} {Compiler-integrated, conversational ai for debugging cs1 programs}.{\BBCQ}
\newblock
\BIn{} \APACrefbtitle {Proceedings of the 56th ACM Technical Symposium on Computer Science Education V. 1} {Proceedings of the 56th acm technical symposium on computer science education v. 1}\ (\BPG~994–1000).
\newblock
\APACaddressPublisher{New York, NY, USA}{Association for Computing Machinery}.
\newblock
\begin{APACrefDOI} \doi{10.1145/3641554.3701827} \end{APACrefDOI}
\PrintBackRefs{\CurrentBib}

\bibitem [\protect \citeauthoryear {%
Riazi%
\ \BBA {} Rooshenas%
}{%
Riazi%
\ \BBA {} Rooshenas%
}{%
{\protect \APACyear {2025}}%
}]{%
riazi2025llm}
\APACinsertmetastar {%
riazi2025llm}%
\begin{APACrefauthors}%
Riazi, S.%
\BCBT {}\ \BBA {} Rooshenas, P.%
\end{APACrefauthors}%
\unskip\
\newblock
\APACrefYearMonthDay{2025}{}{}.
\newblock
{\BBOQ}\APACrefatitle {LLM-Driven Feedback for Enhancing Conceptual Design Learning in Database Systems Courses} {Llm-driven feedback for enhancing conceptual design learning in database systems courses}.{\BBCQ}
\newblock
\BIn{} \APACrefbtitle {Proceedings of the 56th ACM Technical Symposium on Computer Science Education V. 1} {Proceedings of the 56th acm technical symposium on computer science education v. 1}\ (\BPG~1001–1007).
\newblock
\APACaddressPublisher{New York, NY, USA}{Association for Computing Machinery}.
\newblock
\begin{APACrefDOI} \doi{10.1145/3641554.3701940} \end{APACrefDOI}
\PrintBackRefs{\CurrentBib}

\bibitem [\protect \citeauthoryear {%
Roest%
, Keuning%
\BCBL {}\ \BBA {} Jeuring%
}{%
Roest%
\ \protect \BOthers {.}}{%
{\protect \APACyear {2023}}%
}]{%
roest2023nextstep}
\APACinsertmetastar {%
roest2023nextstep}%
\begin{APACrefauthors}%
Roest, L.%
, Keuning, H.%
\BCBL {}\ \BBA {} Jeuring, J.%
\end{APACrefauthors}%
\unskip\
\newblock
\APACrefYearMonthDay{2023}{}{}.
\newblock
{\BBOQ}\APACrefatitle {Next-{{Step Hint Generation}} for {{Introductory Programming Using Large Language Models}}} {Next-{{Step Hint Generation}} for {{Introductory Programming Using Large Language Models}}}.{\BBCQ}
\newblock
\BIn{} \APACrefbtitle {Proceedings of the 26th {{Australasian Computing Education Conference}}} {Proceedings of the 26th {{Australasian Computing Education Conference}}}\ (\BPGS\ 144--153).
\newblock
\APACaddressPublisher{Sydney, Australia}{ACM}.
\newblock
\begin{APACrefDOI} \doi{10.1145/3636243.3636259} \end{APACrefDOI}
\PrintBackRefs{\CurrentBib}

\bibitem [\protect \citeauthoryear {%
Roethlisberger%
\ \BBA {} Dickson%
}{%
Roethlisberger%
\ \BBA {} Dickson%
}{%
{\protect \APACyear {1939}}%
}]{%
roethlisberger1939management}
\APACinsertmetastar {%
roethlisberger1939management}%
\begin{APACrefauthors}%
Roethlisberger, F\BPBI J.%
\BCBT {}\ \BBA {} Dickson, W\BPBI J.%
\end{APACrefauthors}%
\unskip\
\newblock
\APACrefYear{1939}.
\newblock
\APACrefbtitle {{Management and the Worker}} {{Management and the Worker}}.
\newblock
\APACaddressPublisher{Cambridge}{Harvard University Press}.
\PrintBackRefs{\CurrentBib}

\bibitem [\protect \citeauthoryear {%
Sarsa%
, Denny%
, Hellas%
\BCBL {}\ \BBA {} Leinonen%
}{%
Sarsa%
\ \protect \BOthers {.}}{%
{\protect \APACyear {2022}}%
}]{%
Sarsa2022}
\APACinsertmetastar {%
Sarsa2022}%
\begin{APACrefauthors}%
Sarsa, S.%
, Denny, P.%
, Hellas, A.%
\BCBL {}\ \BBA {} Leinonen, J.%
\end{APACrefauthors}%
\unskip\
\newblock
\APACrefYearMonthDay{2022}{{\APACmonth{08}}}{}.
\newblock
{\BBOQ}\APACrefatitle {Automatic Generation of Programming Exercises and Code Explanations Using Large Language Models} {Automatic generation of programming exercises and code explanations using large language models}.{\BBCQ}
\newblock
\BIn{} \APACrefbtitle {Proc. ICER.} {Proc. icer.}
\newblock
\APACaddressPublisher{}{ACM}.
\newblock
\begin{APACrefDOI} \doi{10.1145/3501385.3543957} \end{APACrefDOI}
\PrintBackRefs{\CurrentBib}

\bibitem [\protect \citeauthoryear {%
Savelka%
, Agarwal%
, Bogart%
\BCBL {}\ \BBA {} Sakr%
}{%
Savelka%
\ \protect \BOthers {.}}{%
{\protect \APACyear {2023}}%
}]{%
savelka2023large}
\APACinsertmetastar {%
savelka2023large}%
\begin{APACrefauthors}%
Savelka, J.%
, Agarwal, A.%
, Bogart, C.%
\BCBL {}\ \BBA {} Sakr, M.%
\end{APACrefauthors}%
\unskip\
\newblock
\APACrefYearMonthDay{2023}{}{}.
\newblock
\APACrefbtitle {Large Language Models (GPT) Struggle to Answer Multiple-Choice Questions about Code.} {Large language models (gpt) struggle to answer multiple-choice questions about code.}
\PrintBackRefs{\CurrentBib}

\bibitem [\protect \citeauthoryear {%
Scholl%
\ \BBA {} Kiesler%
}{%
Scholl%
\ \BBA {} Kiesler%
}{%
{\protect \APACyear {2024}}%
}]{%
scholl2024hownovice}
\APACinsertmetastar {%
scholl2024hownovice}%
\begin{APACrefauthors}%
Scholl, A.%
\BCBT {}\ \BBA {} Kiesler, N.%
\end{APACrefauthors}%
\unskip\
\newblock
\APACrefYearMonthDay{2024}{}{}.
\newblock
{\BBOQ}\APACrefatitle {How Novice Programmers Use and Experience ChatGPT when Solving Programming Exercises in an Introductory Course} {How novice programmers use and experience chatgpt when solving programming exercises in an introductory course}.{\BBCQ}
\newblock
\BIn{} \APACrefbtitle {2024 IEEE Frontiers in Education Conference (FIE)} {2024 ieee frontiers in education conference (fie)}\ (\BPG~1-9).
\newblock
\begin{APACrefDOI} \doi{10.1109/FIE61694.2024.10893442} \end{APACrefDOI}
\PrintBackRefs{\CurrentBib}

\bibitem [\protect \citeauthoryear {%
Scholl%
\ \BBA {} Kiesler%
}{%
Scholl%
\ \BBA {} Kiesler%
}{%
{\protect \APACyear {2025}}%
{\protect \APACexlab {{\protect \BCnt {1}}}}}]{%
Scholl_Kiesler_2025_datastudents}
\APACinsertmetastar {%
Scholl_Kiesler_2025_datastudents}%
\begin{APACrefauthors}%
Scholl, A.%
\BCBT {}\ \BBA {} Kiesler, N.%
\end{APACrefauthors}%
\unskip\
\newblock
\APACrefYearMonthDay{2025{\protect \BCnt {1}}}{Jul}{}.
\newblock
\APACrefbtitle {Data: Students’ Feedback Requests and Interactions with the SCRIPT Chatbot - Do They Get What They Ask For?} {Data: Students’ feedback requests and interactions with the script chatbot - do they get what they ask for?}
\newblock
\APACaddressPublisher{}{OSF}.
\newblock
\begin{APACrefDOI} \doi{10.17605/OSF.IO/TG5R3} \end{APACrefDOI}
\PrintBackRefs{\CurrentBib}

\bibitem [\protect \citeauthoryear {%
Scholl%
\ \BBA {} Kiesler%
}{%
Scholl%
\ \BBA {} Kiesler%
}{%
{\protect \APACyear {2025}}%
{\protect \APACexlab {{\protect \BCnt {2}}}}}]{%
scholl2025script}
\APACinsertmetastar {%
scholl2025script}%
\begin{APACrefauthors}%
Scholl, A.%
\BCBT {}\ \BBA {} Kiesler, N.%
\end{APACrefauthors}%
\unskip\
\newblock
\APACrefYearMonthDay{2025{\protect \BCnt {2}}}{}{}.
\newblock
{\BBOQ}\APACrefatitle {SCRIPT - Supportive Chatbot for Resolving Introductory Programming Tasks} {Script - supportive chatbot for resolving introductory programming tasks}.{\BBCQ}
\newblock
\BIn{} \APACrefbtitle {Proceedings of the 30th ACM Conference on Innovation and Technology in Computer Science Education V. 2} {Proceedings of the 30th acm conference on innovation and technology in computer science education v. 2}\ (\BPG~759).
\newblock
\APACaddressPublisher{New York, NY, USA}{Association for Computing Machinery}.
\newblock
\begin{APACrefDOI} \doi{10.1145/3724389.3730786} \end{APACrefDOI}
\PrintBackRefs{\CurrentBib}

\bibitem [\protect \citeauthoryear {%
Scholl%
, Schiffner%
\BCBL {}\ \BBA {} Kiesler%
}{%
Scholl%
\ \protect \BOthers {.}}{%
{\protect \APACyear {2024}}%
}]{%
scholl2024analyzing}
\APACinsertmetastar {%
scholl2024analyzing}%
\begin{APACrefauthors}%
Scholl, A.%
, Schiffner, D.%
\BCBL {}\ \BBA {} Kiesler, N.%
\end{APACrefauthors}%
\unskip\
\newblock
\APACrefYearMonthDay{2024}{}{}.
\newblock
{\BBOQ}\APACrefatitle {Analyzing Chat Protocols of Novice Programmers Solving Introductory Programming Tasks with ChatGPT} {Analyzing chat protocols of novice programmers solving introductory programming tasks with chatgpt}.{\BBCQ}
\newblock
\BIn{} \APACrefbtitle {Proc. DELFI 2024} {Proc. delfi 2024}\ (\BPGS\ 63--79).
\newblock
\begin{APACrefDOI} \doi{10.18420/delfi2024_05} \end{APACrefDOI}
\PrintBackRefs{\CurrentBib}

\bibitem [\protect \citeauthoryear {%
Spohrer%
\ \BBA {} Soloway%
}{%
Spohrer%
\ \BBA {} Soloway%
}{%
{\protect \APACyear {1986}}%
}]{%
spohrer1986novice}
\APACinsertmetastar {%
spohrer1986novice}%
\begin{APACrefauthors}%
Spohrer, J\BPBI C.%
\BCBT {}\ \BBA {} Soloway, E.%
\end{APACrefauthors}%
\unskip\
\newblock
\APACrefYearMonthDay{1986}{}{}.
\newblock
{\BBOQ}\APACrefatitle {{Novice mistakes: Are the folk wisdoms correct?}} {{Novice mistakes: Are the folk wisdoms correct?}}{\BBCQ}
\newblock
\APACjournalVolNumPages{{Communications of the ACM}}{29}{7}{624--632}.
\newblock
\begin{APACrefDOI} \doi{10.1145/6138.6145} \end{APACrefDOI}
\PrintBackRefs{\CurrentBib}

\bibitem [\protect \citeauthoryear {%
Szabo%
\ \protect \BOthers {.}}{%
Szabo%
\ \protect \BOthers {.}}{%
{\protect \APACyear {2025}}%
}]{%
szabo2025modelsofmastery}
\APACinsertmetastar {%
szabo2025modelsofmastery}%
\begin{APACrefauthors}%
Szabo, C.%
, Parker, M\BPBI C.%
, Friend, M.%
, Jeuring, J.%
, Kohn, T.%
, Malmi, L.%
\BCBL {}\ \BBA {} Sheard, J.%
\end{APACrefauthors}%
\unskip\
\newblock
\APACrefYearMonthDay{2025}{}{}.
\newblock
{\BBOQ}\APACrefatitle {Models of Mastery Learning for Computing Education} {Models of mastery learning for computing education}.{\BBCQ}
\newblock
\BIn{} \APACrefbtitle {Proceedings of the 56th ACM Technical Symposium on Computer Science Education V. 1} {Proceedings of the 56th acm technical symposium on computer science education v. 1}\ (\BPG~1092–1098).
\newblock
\APACaddressPublisher{New York, NY, USA}{Association for Computing Machinery}.
\newblock
\begin{APACrefDOI} \doi{10.1145/3641554.3701868} \end{APACrefDOI}
\PrintBackRefs{\CurrentBib}

\bibitem [\protect \citeauthoryear {%
Taylor%
, Vassar%
, Renzella%
\BCBL {}\ \BBA {} Pearce%
}{%
Taylor%
\ \protect \BOthers {.}}{%
{\protect \APACyear {2024}}%
}]{%
taylor2024dcchelperrorexplanations}
\APACinsertmetastar {%
taylor2024dcchelperrorexplanations}%
\begin{APACrefauthors}%
Taylor, A.%
, Vassar, A.%
, Renzella, J.%
\BCBL {}\ \BBA {} Pearce, H.%
\end{APACrefauthors}%
\unskip\
\newblock
\APACrefYearMonthDay{2024}{}{}.
\newblock
{\BBOQ}\APACrefatitle {dcc --help: Transforming the Role of the Compiler by Generating Context-Aware Error Explanations with Large Language Models} {dcc --help: Transforming the role of the compiler by generating context-aware error explanations with large language models}.{\BBCQ}
\newblock
\BIn{} \APACrefbtitle {Proceedings of the 55th ACM Technical Symposium on Computer Science Education V. 1} {Proceedings of the 55th acm technical symposium on computer science education v. 1}\ (\BPG~1314–1320).
\newblock
\APACaddressPublisher{New York}{ACM}.
\newblock
\begin{APACrefDOI} \doi{10.1145/3626252.3630822} \end{APACrefDOI}
\PrintBackRefs{\CurrentBib}

\bibitem [\protect \citeauthoryear {%
Wermelinger%
}{%
Wermelinger%
}{%
{\protect \APACyear {2023}}%
}]{%
wermelinger2023using}
\APACinsertmetastar {%
wermelinger2023using}%
\begin{APACrefauthors}%
Wermelinger, M.%
\end{APACrefauthors}%
\unskip\
\newblock
\APACrefYearMonthDay{2023}{}{}.
\newblock
{\BBOQ}\APACrefatitle {Using GitHub Copilot to Solve Simple Programming Problems} {Using github copilot to solve simple programming problems}.{\BBCQ}
\newblock
\BIn{} \APACrefbtitle {Proceedings of the 54th ACM Technical Symposium on Computer Science Education V. 1} {Proceedings of the 54th acm technical symposium on computer science education v. 1}\ (\BPG~172–178).
\newblock
\APACaddressPublisher{New York, NY, USA}{Association for Computing Machinery}.
\newblock
\begin{APACrefDOI} \doi{10.1145/3545945.3569830} \end{APACrefDOI}
\PrintBackRefs{\CurrentBib}

\bibitem [\protect \citeauthoryear {%
Whalley%
, Clear%
\BCBL {}\ \BBA {} Lister%
}{%
Whalley%
\ \protect \BOthers {.}}{%
{\protect \APACyear {2007}}%
}]{%
whalley2007many}
\APACinsertmetastar {%
whalley2007many}%
\begin{APACrefauthors}%
Whalley, J.%
, Clear, T.%
\BCBL {}\ \BBA {} Lister, R.%
\end{APACrefauthors}%
\unskip\
\newblock
\APACrefYearMonthDay{2007}{}{}.
\newblock
{\BBOQ}\APACrefatitle {{The many ways of the Bracelet project}} {{The many ways of the Bracelet project}}.{\BBCQ}
\newblock
\APACjournalVolNumPages{{BACIT}}{}{}{}.
\PrintBackRefs{\CurrentBib}

\bibitem [\protect \citeauthoryear {%
Xiao%
, Hou%
\BCBL {}\ \BBA {} Stamper%
}{%
Xiao%
\ \protect \BOthers {.}}{%
{\protect \APACyear {2024}}%
}]{%
xiao2024exploring}
\APACinsertmetastar {%
xiao2024exploring}%
\begin{APACrefauthors}%
Xiao, R.%
, Hou, X.%
\BCBL {}\ \BBA {} Stamper, J.%
\end{APACrefauthors}%
\unskip\
\newblock
\APACrefYearMonthDay{2024}{}{}.
\newblock
{\BBOQ}\APACrefatitle {Exploring How Multiple Levels of GPT-Generated Programming Hints Support or Disappoint Novices} {Exploring how multiple levels of gpt-generated programming hints support or disappoint novices}.{\BBCQ}
\newblock
\BIn{} \APACrefbtitle {Extended Abstracts of the 2024 CHI Conference on Human Factors in Computing Systems.} {Extended abstracts of the 2024 chi conference on human factors in computing systems.}
\newblock
\APACaddressPublisher{New York, USA}{ACM}.
\newblock
\begin{APACrefDOI} \doi{10.1145/3613905.3650937} \end{APACrefDOI}
\PrintBackRefs{\CurrentBib}

\bibitem [\protect \citeauthoryear {%
Yeh%
\ \protect \BOthers {.}}{%
Yeh%
\ \protect \BOthers {.}}{%
{\protect \APACyear {2025}}%
}]{%
yeh2025bridging}
\APACinsertmetastar {%
yeh2025bridging}%
\begin{APACrefauthors}%
Yeh, T\BPBI Y.%
, Tran, K.%
, Gao, G.%
, Yu, T.%
, Fong, W\BPBI O.%
\BCBL {}\ \BBA {} Chen, T\BHBI Y.%
\end{APACrefauthors}%
\unskip\
\newblock
\APACrefYearMonthDay{2025}{}{}.
\newblock
{\BBOQ}\APACrefatitle {Bridging Novice Programmers and LLMs with Interactivity} {Bridging novice programmers and llms with interactivity}.{\BBCQ}
\newblock
\BIn{} \APACrefbtitle {Proceedings of the 56th ACM Technical Symposium on Computer Science Education V. 1} {Proceedings of the 56th acm technical symposium on computer science education v. 1}\ (\BPG~1295–1301).
\newblock
\APACaddressPublisher{New York, NY, USA}{Association for Computing Machinery}.
\newblock
\begin{APACrefDOI} \doi{10.1145/3641554.3701867} \end{APACrefDOI}
\PrintBackRefs{\CurrentBib}

\bibitem [\protect \citeauthoryear {%
Zhai%
}{%
Zhai%
}{%
{\protect \APACyear {2022}}%
}]{%
zhai2022chatgpt}
\APACinsertmetastar {%
zhai2022chatgpt}%
\begin{APACrefauthors}%
Zhai, X.%
\end{APACrefauthors}%
\unskip\
\newblock
\APACrefYearMonthDay{2022}{}{}.
\newblock
{\BBOQ}\APACrefatitle {ChatGPT User Experience: Implications for Education} {Chatgpt user experience: Implications for education}.{\BBCQ}
\newblock

\newblock
\begin{APACrefDOI} \doi{http://dx.doi.org/10.2139/ssrn.4312418} \end{APACrefDOI}
\PrintBackRefs{\CurrentBib}

\bibitem [\protect \citeauthoryear {%
Zhang%
\ \protect \BOthers {.}}{%
Zhang%
\ \protect \BOthers {.}}{%
{\protect \APACyear {2022}}%
}]{%
zhang2022repairing}
\APACinsertmetastar {%
zhang2022repairing}%
\begin{APACrefauthors}%
Zhang, J.%
, Cambronero, J.%
, Gulwani, S.%
, Le, V.%
, Piskac, R.%
, Soares, G.%
\BCBL {}\ \BBA {} Verbruggen, G.%
\end{APACrefauthors}%
\unskip\
\newblock
\APACrefYearMonthDay{2022}{}{}.
\newblock
{\BBOQ}\APACrefatitle {Repairing Bugs in Python Assignments Using Large Language Models} {Repairing bugs in python assignments using large language models}.{\BBCQ}
\newblock
\APACjournalVolNumPages{arXiv preprint arXiv:2209.14876}{}{}{}.
\PrintBackRefs{\CurrentBib}

\end{thebibliography}
\bibliographystyle{apacite} 
\end{document}